\def\year{2022}\relax
\documentclass[letterpaper]{article} 
\usepackage{aaai22}  
\usepackage{times}  
\usepackage{helvet}  
\usepackage{courier}  
\usepackage[hyphens]{url}  
\usepackage{graphicx} 
\usepackage{natbib}  
\usepackage{caption} 
\DeclareCaptionStyle{ruled}{labelfont=normalfont,labelsep=colon,strut=off} 
\usepackage{algorithm}
\usepackage{algorithmic}
\usepackage{amsmath}

\usepackage{amsfonts,amssymb}
\usepackage[most]{tcolorbox}
\definecolor{black}{rgb}{0.0, 0.0, 0.0}
\definecolor{wine}{rgb}{0.5333333333333333, 0.13333333333333333, 0.3333333333333333}
\newcommand{\fcircle}[2][red,fill=red]{\tikz[baseline=-0.5ex]\draw[#1,radius=#2] (0,0.03) circle ;}

\usepackage{ntheorem}
\newtheorem{prop}{Proposition}
\newtheorem{thm}{Theorem}

\usepackage{newfloat}
\usepackage{listings}
\floatstyle{ruled}
\newfloat{listing}{tb}{lst}{}
\floatname{listing}{Listing}
\title{Neural Piecewise-Constant Delay Differential Equations}

\author{Qunxi Zhu\textsuperscript{\rm 1}\thanks{This work was done when the first two authors were interns at Microsoft Research Asia.  To whom correspondence should be addressed: Q.Z. and W.L, \text{https://faculty.fudan.edu.cn/wlin/zh\_CN/index.htm}.}
,  Yifei Shen\textsuperscript{\rm 2}\footnotemark[1] , 
Dongsheng Li\textsuperscript{\rm 3},  
Wei Lin\textsuperscript{\rm 4}  \\
}

\affiliations{
    \textsuperscript{\rm 1} Research Institute of Intelligent Complex Systems, Fudan University\\
    \textsuperscript{\rm 2} Department of Electronic and Computer Engineering, The Hong Kong University of Science and Technology, Hong Kong\\
    \textsuperscript{\rm 3} Microsoft Research Asia\\
    
    \textsuperscript{\rm 4} School of Mathematical Sciences, State Key Laboratory of Medical Neurobiology and MOE Frontiers Center for Brain Science, Institutes of Brain Science, Shanghai Center for Mathematical Sciences, Center for Computational Systems Biology, LCNBI, and Research Institute of Intelligent Complex Systems, Fudan University
    
    \texttt{\{qxzhu16,wlin\}@fudan.edu.cn}, \texttt{yshenaw@connect.ust.hk}, \texttt{dongsheng.li@microsoft.com}

}

\usepackage{bibentry}

\begin{document}

\maketitle

\begin{abstract}
    Continuous-depth neural networks, such as the Neural Ordinary Differential Equations (ODEs), have aroused a great deal of interest from the communities of machine learning and data science in recent years, which bridge the connection between deep neural networks and dynamical systems.  In this article, we introduce a new sort of continuous-depth neural network, called the Neural Piecewise-Constant Delay Differential Equations (PCDDEs).  Here, unlike the recently proposed framework of the Neural Delay Differential Equations (DDEs), we transform the single delay into the piecewise-constant delay(s). The Neural PCDDEs with such a transformation, on one hand, inherit the strength of universal approximating capability in Neural DDEs.  On the other hand, the Neural PCDDEs, leveraging the contributions of the information from the multiple previous time steps, further promote the modeling capability without augmenting the network dimension. With such a promotion, we show that the Neural PCDDEs do outperform the several existing continuous-depth neural frameworks on the one-dimensional piecewise-constant delay population dynamics and real-world datasets, including MNIST, CIFAR10, and SVHN.
\end{abstract}

\noindent Recently, many frameworks have been established, connecting dynamical systems tightly with neural networks and promoting the network performances significantly \citep{weinan2017proposal, li2017maximum, haber2017stable, chang2017multi, 2018Model, 2018Brain, li2018optimal, lu2018beyond, weinan2019mean, chang2019antisymmetricrnn, ruthotto2019deep, zhang2019you,  Qunxi2019Detecting, tang2020introduction}.  One framework of  milestone is the Neural Ordinary Differential Equations (NODEs), also regarded as continuous-depth neural networks \citep{chen2018neural}. The framework non-trivially extends the traditional residual neural networks (ResNets) \citep{he2016deep} to parametric ordinary differential equations (ODEs), where the time in NODEs is treated as the ``depth''  of the ResNets. Actually,  different from the traditional neural networks, the NODEs model the vector fields of the ODEs through optimizing their parameters with the back-propagation algorithm and the ODE solver based on the training data and a predefined loss function. In addition, the NODEs contain a broad range of architectures, including the feed-forward neural networks and the convolution neural networks.

Owing to the constant memory cost, the continuous dynamical behavior, and the naturally-rooted invertibility of the NODEs, applications of such a framework to modeling physical systems are growing.  Examples abound: data analytics on the  time series with irregular sampling duration \citep{NEURIPS2019_42a6845a, NEURIPS2019_455cb265, kidger2020neural}, generations of the continuous normalizing flow \citep{chen2018neural, grathwohl2018ffjord, finlay2020train, deng2020modeling, kelly2020learning}, and representations of the point clouds \citep{yang2019pointflow, rempe2020caspr}. It is worthwhile to mention that the framework of NODEs, in spite of its wide applicability, is not of a universal approximator.  It thus cannot successfully learn representative maps such as the \textit{reflections} or the \textit{concentric annuli} due to the homeomorphism property of ODEs \citep{dupont2019augmented, zhang2020approximation}. To address this problem, several practical schemes \citep{dupont2019augmented, massaroli2020dissecting, zhu2021neural} have been suggested and implemented, among which the Neural Delay Differential Equations (NDDEs) show an outstanding efficacy in approximating functionals based on given data \citep{zhu2021neural}. Additionally, variants of extensions and applications of the NODEs have been proposed in recent years, including the partial differential equations \citep{han2018solving, ruthotto2019deep, sun2020neupde} and the stochastic differential equations \citep{liu2019neural, jia2019neural, li2020scalable, song2020score}.

In this article, inspired by a recent framework of NDDEs, as mentioned above,  we develop a new framework of continuous-depth neural networks with different configurations of delay(s). Although the NDDEs not only allow the trajectories to intersect with each other even in a lower-dimensional phase space but also accurately model representative delayed physical/biological systems, such as the Mackey-Glass system \citep{Mackey1977Oscillation}, they likely suffer from tremendously high  computational cost.  Clearly, such a shortcoming is due to the persistent existence of the effect induced by $t-\tau$, the dynamical delay. In addition, continuously improving the feature representation of a neural architecture is also a challenging direction of machine learning. To conquer these difficulties, we therefore propose the model as mentioned above. Indeed, the proposed model mainly consists of the following two configurations: (1) novelly transforming the dynamical delay in NDDEs into a piecewise-constant delay, viz. $\lfloor t - \tau \rfloor$, and (2) introducing multiple piecewise-constant delays into the vector field to significantly promote the feature representation. 

The advantages of the first configuration include preserving the computational efficacy with the simple piecewise-constant delay and maintaining the capabilities of modeling complex dynamics (chaos) using the discontinuous nature of the particular form of this delay. The advantage of the second configuration involves leveraging the information from not only the current time point but also many previous time points and thus strengthening the feature propagation. All these advantages definitely result in a better feature representation, comparing to the NDDEs.
Mathematically, our model is originated from a well-developed class of delay differential equations, called the piecewise-constant delay differential equations (PCDDEs) \citep{1988A, cooke1991survey, 1992On}. We therefore refer our model of continuous-depth neural networks to the neural PCDDEs (NPCDDEs).
To further improve the performance, we propose an extension of the NPCDDEs without sharing the parameters in different time duration, called the unshared NPCDDEs (UNPCDDEs).

To summarize, the major contributions of this article are multi-folded, including:
\begin{itemize}
    \item establishment of a generic continuous-depth model, NPCDDEs, such that typical neural networks, such as ResNets and NODEs, are the special cases of the UNPCDDEs, 
    \item validation of the NPCDDEs having the capability of universal approximation (see Proposition~\ref{prop2}),
    \item formulation of the adjoint dynamical system and the backward gradients for the NPCDDEs (see Theorem~\ref{thm2}), and
    \item demonstrations of the powerful nonlinear representation of NPCDDEs on the synthetic data produced by the one-dimensional piecewise-constant delay population dynamics and on the representative image datasets, i.e., MNIST, CIFAR10, and SVHN, as well.
\end{itemize}

\section{Related Works}
\label{sec2}

\paragraph{\fcircle[fill=black]{3pt} Neural Ordinary Differential Equations}
As pointed out by \citep{chen2018neural}, the NODEs can be regarded as the continuous version of the ResNets having an infinite number of layers \citep{he2016deep}. The residual block of the ResNets is mathematically written as $\mathbf{z}_{t+1} = \mathbf{z}_t + f(\mathbf{z}_t, \theta_t)$, where $\mathbf{z}_t$ is the feature at the $t$-th layer, and $f(\cdot, \cdot)$ is a dimension-preserving and nonlinear function parametrized by a neural network with $\theta_t$, the parameter vector pending for learning.  Notably, such a transformation could be viewed as the special case of the following discrete-time equations:
\begin{equation}
\label{eq1}
    \frac{\mathbf{z}_{t+1} - \mathbf{z}_t }{\Delta t} = f(\mathbf{z}_t, \theta_t)
\end{equation}
with $\Delta t=1$. In other words, as $\Delta t$ in \eqref{eq1} is set as an infinitesimal increment, the ResNets could be regarded as the Euler discretization of the
NODEs which read: 
\begin{equation}
\label{eq2}
    \frac{d\mathbf{z}(t) }{d t} = f(\mathbf{z}(t), \theta). 
\end{equation}
Here, the shared parameter vector $\theta$, which unifies the vector $\theta_t$ of every layer in Eq.~\eqref{eq1}, is injected into the vector field across the finite time horizon, to achieve parameter efficiency of the NODEs.  As such, the NODEs can be used to approximate some unknown function $F: \mathbf{x}\mapsto F(\mathbf{x})$.  Specifically, the approximation is achieved in the following manner: Constructing a flow of the NODEs starting from the initial state $\mathbf{z}(0)=\mathbf{x}$ and ending at the final state $\mathbf{z}(T)$ with $\mathbf{z}(T) \approx F(\mathbf{x})$.  
Thus, a standard framework of the NODEs, which takes the input as its initial state and the feature representation as the final state, is formulated as:
\begin{equation}
\label{eq3}
    \left\{
    \begin{aligned}
        \mathbf{z}(T) & = \mathbf{z}(0) + \int_0^T f(\mathbf{z}(t), \theta) dt \\
        & = \mbox{ODESolve}(\mathbf{z}(0), f,0,T,\theta), \\
        \mathbf{z}(0) & = \mbox{input},
    \end{aligned}
    \right.
\end{equation}
where $T$ is the final time and the solution of the above ODE can be numerically obtained by the standard ODE solver using adaptive schemes.  Indeed, a supervised learning task can be formulated as:
\begin{equation}
\label{eq4}
    \begin{array}{c}
        \min_{\theta} L(\mathbf{z}(T)),\\
        \mbox{~s.t. Eq.~(\ref{eq2}) holds for any~} t \in[0, T],
    \end{array}
\end{equation}
where $L(\cdot)$ is a predefined loss function. To optimize the loss function in \eqref{eq4}, we need to calculate the gradient with respect to the parameter vector.  This calculation can be implemented with a memory in an order of $\mathcal{O}(1)$ by employing the adjoint sensitivity method \citep{chen2018neural, pontryagin1962mathematical} as:
\begin{equation}
    \frac{dL}{d\theta} = -\int_{T}^0 \mathbf{a}(t)^{\top} \frac{\partial f(\mathbf{z}(t),\theta)}{\partial \theta} dt, 
\end{equation}
where $\mathbf{a}(t):=\frac{\partial L}{\partial \mathbf{z}(t)}$ is called the \textit{adjoint}, representing the gradient with respect to the hidden states $\mathbf{z}(t)$ at each time point $t$.

\noindent
\paragraph{\fcircle[fill=black]{3pt} Variants of NODEs} As shown in \citep{dupont2019augmented}, there are still some typical class of functions that the NODEs cannot represent. For instance, the \textit{reflections}, defined by $g_{1{\rm d}}:\mathbb{R}\rightarrow \mathbb{R}$ with $g_{1{\rm d}}(1)=-1$ and $g_{1{\rm d}}(-1)=1$, and the \textit{concentric annuli}, defined by $g_{2{\rm d}}:\mathbb{R}^2\rightarrow \mathbb{R}$ with 
\begin{equation}
\label{eq6}
    g_{2{\rm d}}(\mathbf{x}) = \left\{
    \begin{array}{ll}
         -1, & \mbox{if~} \|\mathbf{x}\|\leq r_1,  \\
         1,  & \mbox{if~} r_2\leq \|\mathbf{x}\|\leq r_3,
    \end{array}
   \right.
\end{equation}
where $\|\cdot\|$ is the $L_2$ norm, and $0<r_1<r_2<r_3$.  Such successful constructions of the two counterexamples are attributed to the fact that the feature mapping from the input (i.e., the initial state) to the features (i.e., the final state) by the NODEs is a homeomorphism. Thus, the features always preserve the topology of the input domain, which mathematically results in the impossibility of separating the two connected regions in \eqref{eq6}.  A few practical strategies have been timely proposed to address this problem.  For example, proposed creatively in \citep{dupont2019augmented} was an argumentation of the input domain into a higher dimensional space, which makes it possible to have more complicated dynamics emergent in the Augmented NODEs.  Very recently, articulated in
\citep{zhu2021neural} was a novel framework of the NDDEs to address this issue without argumentation.  Actually, such a framework was inspired by a broader class of functional differential equations, named delay differential equations (DDEs), where a time delay was introduced \citep{erneux2009applied}. 
Fox example, a simple form of NDDEs reads:
\begin{equation}
\label{eqNDDEold}
\left\{
    \begin{aligned}
        \frac{d\mathbf{z}(t) }{d t} &= f(\mathbf{z}(t-\tau), \theta), ~t\in[0, T],\\
        \mathbf{z}(t)&=\phi(t)=\mathbf{x}, ~t\in[-\tau, 0], 
    \end{aligned}
\right.
\end{equation}
where $\tau$ is the delay effect and 
$\phi(t)$ is the initial function. Hereafter, we assume $\phi(t)$ as a constant function, i.e., $\phi(t)\equiv \mathbf{x}$ with input $\mathbf{x}$.
Due to the infinite-dimension nature of the NDDEs, the crossing orbits can be existent in the lower-dimensional phase space.  More significantly as demonstrated in \cite{zhu2021neural}, the NDDEs have a capability of universal approximation with $T=\tau$ in~\eqref{eqNDDEold}.

{
\noindent
\paragraph{\fcircle[fill=black]{3pt} Control theory} Training a continuous-depth neural network can be regarded  as a task of solving an optimal control problem with a predefined loss function, where the parameters in the network act as the controller \citep{pontryagin1962mathematical, chen2018neural, weinan2019mean}. Thus, developing a new sort of continuous-depth neural network is intrinsic or equivalent to designing an effective controller. Such a controller could be in a form of open-loop or closed-loop. Therefore, from a viewpoint of control, all the existing continuous-depth neural networks can be addressed as control problems. However, these problems require different forms of controllers. Specifically, when we consider the continuous-depth neural network $\frac{dx(t)}{dt} = f(x(t), u(t), t)$, $u(t)$ is regarded as a controller. For example,  $u(t)$ treated as constant parameters yields the network frameworks proposed in \citep{chen2018neural}, $u(t)$ as a data-driven controller yields a framework in \citep{massaroli2020dissecting}, and $u(t)$ as other forms of controllers brings more fruitful network structures  \citep{chalvidal2020go, li2020scalable, kidger2020neural,zhu2021neural}. Here, the mission of this work is to design a delayed feedback controller for rendering a continuous-depth neural network more effectively in coping with synthetic or/and real-world datasets. 
}

\section{Neural Piecewise-Constant Delay Differential Equations}
\label{sec3}
In this section, we propose a new framework of continuous-depth neural networks with delay (i.e., the NPCDDEs) by an articulated integration of some tools from machine learning and dynamical systems: the NDDEs and the piecewise-constant DDEs \citep{1988A, cooke1991survey, 1992On}.

\begin{figure}[htb]
    \begin{center}
        \centerline{\includegraphics[width=0.47\textwidth]{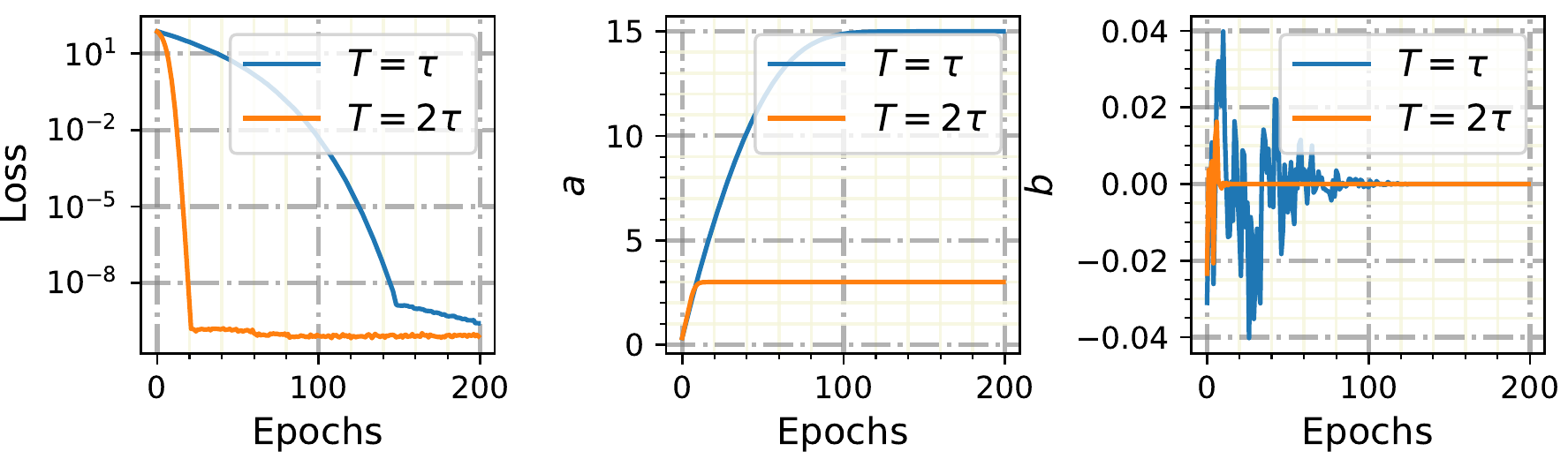}}
        \caption{The training processes for fitting the function $F(x)=16x$ using the NPCDDEs in \eqref{eqNDDE}, respectively, with the final times $T=\tau$ and $T=2\tau$. The training loses (left), and the evolution of two parameters $a$ (middle), and $b$ (right), as defined in \eqref{linearmodel} during the training processes.}
    \label{figpoly}
    \end{center}
    \vskip -0.3in
\end{figure}

We first transform the delay of the NDDEs in \eqref{eqNDDEold} into a form of the piecewise-constant delay \citep{1988A, cooke1991survey, 1992On}, so that we have 
\begin{equation}
\label{eqNDDE}
\left\{
    \begin{aligned}
        \frac{d\mathbf{z}(t) }{d t} &= f(\mathbf{z}(\left\lfloor \frac{t}{\tau} \right\rfloor \tau), \theta), t\in[0, T],\\
        \mathbf{z}(0)&=\mathbf{x}, 
    \end{aligned}
\right.
\end{equation}
where the final time $T=n\tau$ and $n$ is supposed to be a positive integer hereafter. We note that the NPCDDEs in \eqref{eqNDDE} with $T=\tau$ is exactly the NDDEs in~\eqref{eqNDDEold}, owning the universal approximation as mentioned before. As the vector filed of the NPCDDEs in \eqref{eqNDDE} is constant in each interval $[k\tau, k\tau+\tau]$ for $k=0,1,..., \lfloor \frac{T}{\tau} \rfloor$,  the simple NPCDDEs in \eqref{eqNDDE} can be treated as a discrete-time dynamical system:
\begin{equation}
     \mathbf{z}(k+1) = \mathbf{z}(k) + \tau f(\mathbf{z}(k),\theta) := \hat{F}(\mathbf{z}(k),\theta). 
\end{equation}
Actually, this iterative property of dynamical systems enables the NPCDDEs in \eqref{eqNDDE} to learn some functions with specific \textit{structures} more effectively. For example, if the map $F(x)=c^2x$ with a large 
real number $c>0$ is pending for learning and the vector field is set as:
\begin{equation}
\label{linearmodel}
    f(\mathbf{z}( \left\lfloor \frac{t}{\tau} \right\rfloor \tau), \theta) := a \mathbf{z}( \left\lfloor \frac{t}{\tau} \right\rfloor \tau) + b
\end{equation}
with $\tau=1$ and the initial parameters $a=b=0$ before training, then, we only use $T=2\tau$ as the final time for the NPCDDEs in \eqref{eqNDDE} and require $x(\tau)$ to learn the small coefficient in the linear function $x\mapsto cx$ (or, equivalently, require $f$ to learn $x\mapsto (c-1)x$).  As such, the feature $x(T)\approx (x(\tau))^2\approx c^2x$ naturally approximates the above-set function $F(x)$, because $F(x)$ can be simply represented as two iterations of the function $\hat{F}(x)=cx$, i.e., $\hat{F}\circ \hat{F} (x)=F(x)$. We experimentally show the structural representation power in Fig.~\ref{figpoly}, where the training loss of the NPCDDEs in \eqref{eqNDDE} with $T=2\tau$ decreases faster than that only with $T=\tau$.  

Given the above example, the following question arises naturally: For any given function $x\mapsto F(x)$, does there exist a function $x\mapsto \hat{F}(x)$ such that the \textit{functional} equation 
\begin{equation}
    \label{functional_eq}
    \hat{F}\circ \hat{F} (x)=F(x)
\end{equation}
holds? Unfortunately, the answer is no, which is rigorously stated in the following proposition.  

\begin{prop} \label{eq:x^2}
    \citep{Radovanovic2007FunctionalE} There does not exist any function $f:\mathbb{R}\rightarrow \mathbb{R}$ such that $f(f(x)) = x^2-2$ for all $x \in\mathbb{R}$.
\end{prop}
As shown in Proposition \ref{eq:x^2}, although the iterative property of NPCDDEs in \eqref{eqNDDE} allows the effective learning of functions with certain structure, the solution of the functional equation \eqref{functional_eq} does not always exist.
This thus implies that \eqref{eqNDDE} cannot represent a wide class of functions \citep{rice1980f, 2011Solution}.

\begin{figure}[htb]
    \begin{center}
        \centerline{\includegraphics[width=0.45\textwidth]{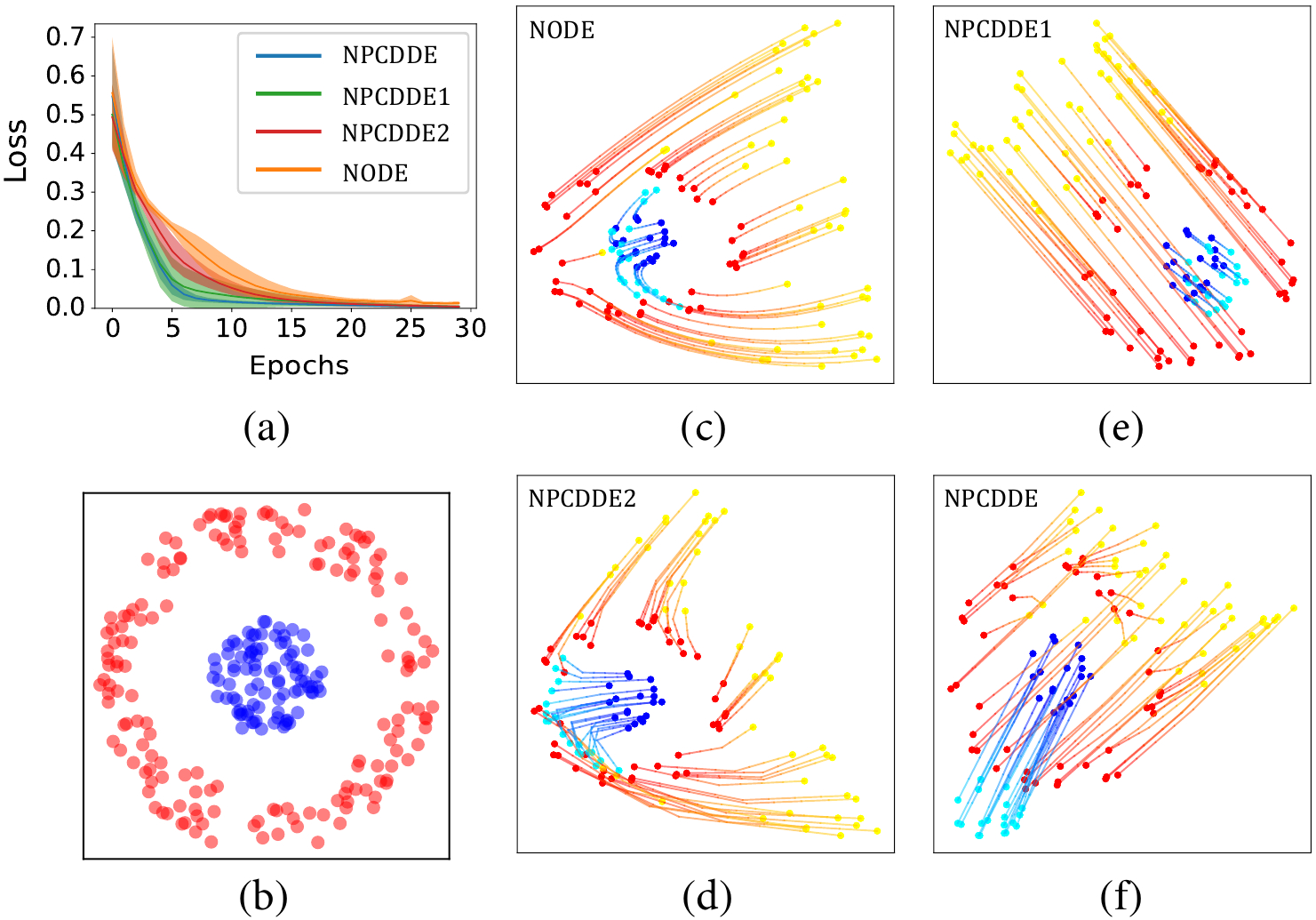}}
        \caption{The training processes for fitting the function $g_{2{\rm d}}(\mathbf{x})$. (a) The training loses, respectively, using the NODEs, the NPCDDEs in \eqref{eqNDDE} with $n=1$ and $\tau=1$, the NPCDDEs in \eqref{eqNDDE} with $n=2$ and $\tau=0.5$, and the special NPCDDEs in \eqref{eqDNDDE} with $\tau=0.5$. (b) A part of the training dataset for visualization. The flows mapping from the initial states to the target states, respectively, by the NODEs (c),  the NPCDDEs in \eqref{eqNDDE} with $n=2$ and $\tau=0.5$ (d), NPCDDEs in \eqref{eqNDDE} with $n=1$ and $\tau=1$ (e), and the special NPCDDEs in \eqref{eqDNDDE} with $\tau=0.5$ (f). The red (resp. blue) points and the yellow (resp. cyan color) points  are the initial states and the final states of all the flows, respectively.}
    \label{figcircle1}
    \end{center}
    \vskip -0.3in
\end{figure}

To further elaborate this point, we use $T=\tau$ and $T=2\tau$, respectively, for the NPCDDEs in \eqref{eqNDDE} to model the function $g_{2{\rm d}}(\mathbf{x})$ as defined in \eqref{eq6}.  Clearly, Fig. \ref{figcircle1} shows that the training processes for fitting the concentric annuli using \eqref{eqNDDE} with the two delays are different.  Contrary to the preceding example, the training loss of one with $T=\tau$ decreases much faster than that of the one with $T=2\tau$. 

\begin{figure}[htb]
    \begin{center}
        \centerline{\includegraphics[width=0.45\textwidth]{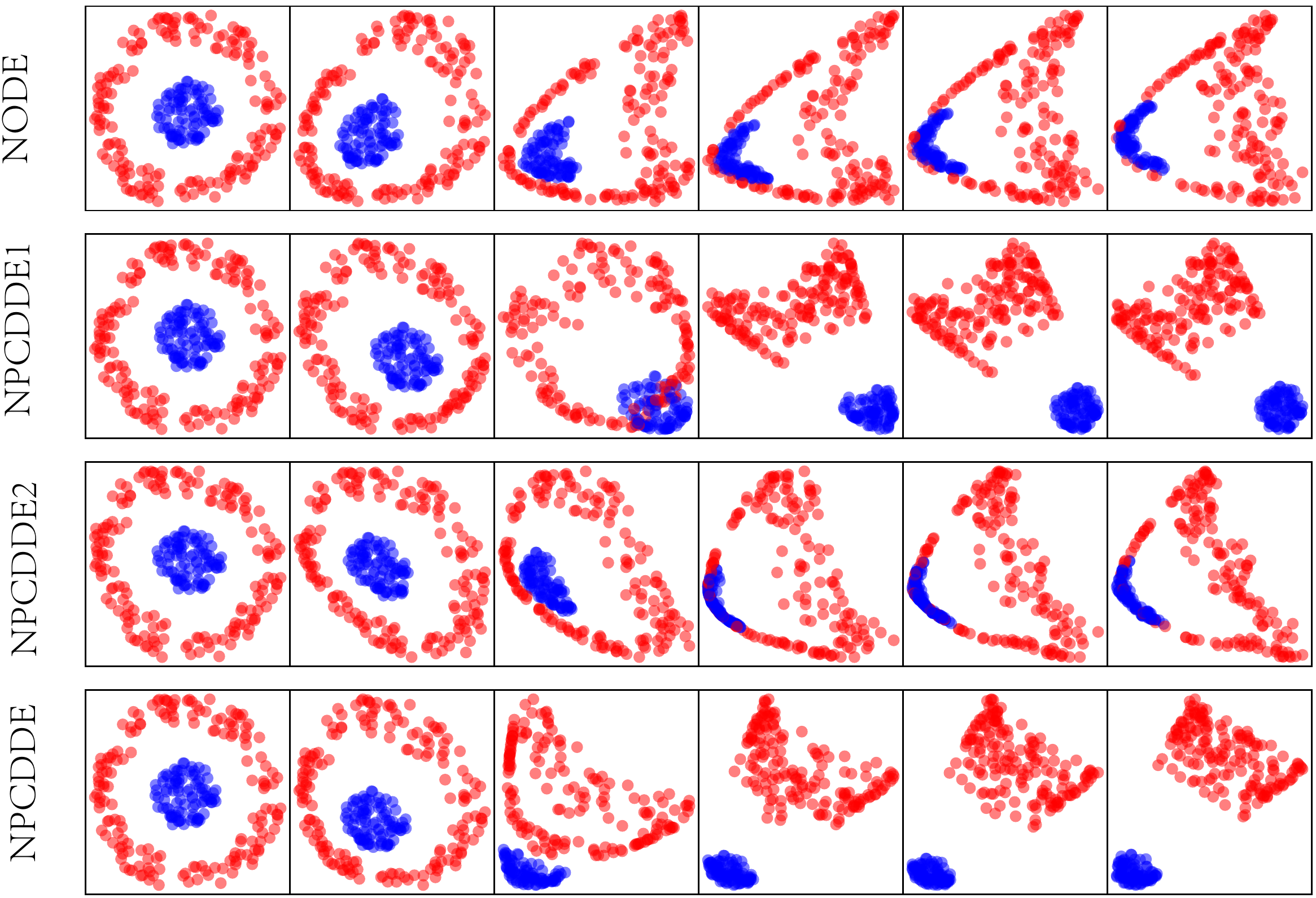}}
        \caption{The dynamical behaviors of the feature spaces during the training processes (totally $6$ epochs from the left column to the right column) for fitting $g_{2{\rm d}}(\mathbf{x})$ using different models: the NODEs (the top row), the NPCDDEs in \eqref{eqNDDE} with $n=1$ and $\tau=1$ (the second row),  the NPCDDEs in \eqref{eqNDDE} with $n=2$ and $\tau=0.5$ (the third row), and the special NPCDDEs in \eqref{eqDNDDE} with $\tau=0.5$ (the bottom row).}
    \label{figcircle2}
    \end{center}
    \vskip -0.3in
\end{figure}

In order to sustain the capability of universal approximation from the NDDEs to the current framework, we modify the NPCDDEs in \eqref{eqNDDE} by adding a skip connection from the time $0$ to the final time $2\tau$ in the following manner:
\begin{equation}
\label{eqDNDDE}
\left\{
    \begin{aligned}
        \frac{d\mathbf{z}(t) }{d t} &= f(\mathbf{z}(\left\lfloor \frac{t}{\tau} \right\rfloor \tau), \mathbf{z}(\left\lfloor \frac{t-\tau}{\tau} \right\rfloor \tau), \theta), ~t\in[0, 2\tau],\\
        \mathbf{z}(-\tau)&=\mathbf{z}(0)=\mathbf{x}. 
    \end{aligned}
\right.
\end{equation}
As can be seen from Fig. \ref{figcircle1}, the training loss of the modified NPCDDEs in \eqref{eqDNDDE}) decreases outstandingly faster than that of the NPCDDEs in \eqref{eqNDDE} with $T=2\tau$ and that of NODEs.  Also, it is slightly faster than the one with $T=\tau$.  Moreover, the dynamical behaviors of the feature spaces during the training processes using different neural frameworks are shown in Fig. \ref{figcircle2}.  In particular, the NPCDDEs in \eqref{eqDNDDE} first separate the two clusters among these models at the $3$rd training epoch, which is beyond the ability of the baselines.


\begin{figure*}[t]
    \begin{center}
        \centerline{\includegraphics[width=0.98\textwidth]{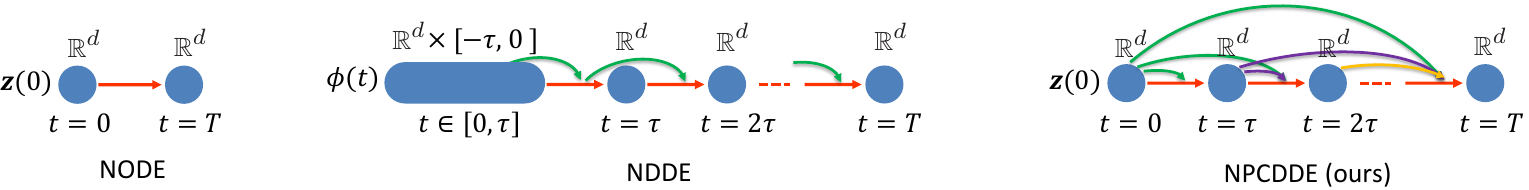}}
        \caption{Sketches of different kinds of continuous-depth neural networks, including the NODEs, the NDDEs, and our newly proposed framework, the NPCDDEs. Specifically, $\phi(t)\equiv \mathbf{z}(0)$ as a constant function is the initial function for the NDDEs. For the NPCDDEs in \eqref{eqDNDDEgenral}, at each time point in the interval $[k\tau, k\tau+\tau]$, the time dependencies are unaltered, different from the dynamical delay in the NDDEs.}
    \label{icml-historical}
    \end{center}
    \vskip -0.2in
\end{figure*}
More importantly, the following theorem demonstrates that NPCDDEs in \eqref{eqDNDDE} are universal approximators, whose proof is provided in the supplementary material.

\begin{thm}
    \label{thm1}
    (Universal approximation of the NPCDDEs in \eqref{eqDNDDE})
    Consider the NPCDDEs in \eqref{eqDNDDE} of $n$-dimension.  If, for any given function $F: \mathbb{R}^n \rightarrow \mathbb{R}^n$, there exists a neural network $g(\mathbf{x}, \theta)$ that can approximate the map $G(\mathbf{x}) = \frac{1}{2\tau} [F(\mathbf{x}) - \mathbf{x}]$, then the NPCDDEs that can learn the map $\mathbf{x} \mapsto F(\mathbf{x})$. In other words, we have $\mathbf{z}(T)\approx F(\mathbf{x})$ provided that both the initial states $\mathbf{z}(-\tau)$ and $\mathbf{z}(0)$ are set as $\mathbf{x}$, the input.
\end{thm}

Notice that, for the NPCDDEs in \eqref{eqNDDE} and the modified NPCDDEs in \eqref{eqDNDDE}, their vector fields keep constant in a $\tau$ period of time.  More generally, we can extend these models by adding the dependency on the current state, enlarging the value of the final time, and introducing more skip connections from the previous time to the current time.  As such, a more generic framework of the NPCDDEs reads:
\begin{equation}
\label{eqDNDDEgenral}
\left\{
    \begin{aligned}
        \frac{d\mathbf{z}(t) }{d t} = & f(\mathbf{z}(t), \mathbf{z}(\left\lfloor \frac{t}{\tau} \right\rfloor \tau), \mathbf{z}(\left\lfloor \frac{t-\tau}{\tau} \right\rfloor \tau), ..., \\
        &\mathbf{z}(\left\lfloor \frac{t-n\tau}{\tau} \right\rfloor \tau), \theta), t\in[0, T],\\
        \mathbf{z}(-n\tau)=&\cdots=\mathbf{z}(-\tau)=\mathbf{z}(0)=\mathbf{x}, 
    \end{aligned}
\right.
\end{equation}
where $T=n\tau$ with $n$ being a positive integer.  Analogous to the proof of Theorem \ref{thm1}, the universal approximation of the NPCDDEs in \eqref{eqDNDDEgenral} can be validated (see Proposition~\ref{prop2}). 
\begin{prop}
    \label{prop2}
    The NPCDDEs in \eqref{eqDNDDEgenral} have a capability of universal approximation. 
\end{prop}

To further improve the  modeling  capability of the NPCDDEs, we propose an extension of the NPCDDEs without sharing the parameters, which reads:
\begin{equation}
\label{eqDNDDEgenralunshared}
\left\{
    \begin{aligned}
        \frac{d\mathbf{z}(t) }{d t} = & f(\mathbf{z}(t), \mathbf{z}(\left\lfloor \frac{t}{\tau} \right\rfloor \tau), \mathbf{z}(\left\lfloor \frac{t-\tau}{\tau} \right\rfloor \tau), ..., \\
        &\mathbf{z}(0), \theta_k), t\in[k\tau, k\tau+\tau],\\
        \mathbf{z}(0)=&\mathbf{x}, 
    \end{aligned}
\right.
\end{equation}
where $\theta_k$ is the parameter vector used in the time interval $[k\tau, k\tau+\tau]$ for $k=0,1,...,n-1$. For simplicity, we name such a model as unshared NPCDDEs (UNPCDDEs).  As in the ResNets~\eqref{eq1}, a typical neural network,  the parameters in each layer are independent with the parameters in the other layer.   
Moreover, the gradients of the loss with respect to the parameters  of the UNPCDDEs in \eqref{eqDNDDEgenralunshared} are shown in Theorem \ref{thm2}, whose proof is provided in the supplementary material.  Moreover, setting $\theta_k\equiv \theta$ straightforwardly in Theorem \ref{thm2} enables us to compute the gradients of the NPCDDEs in \eqref{eqDNDDEgenral}.

\begin{thm} 
    \label{thm2}
    (Backward gradients of the UNPCDDEs in \eqref{eqDNDDEgenralunshared}) Consider the loss function $L(\mathbf{z}(T))$ with the final time $T=n\tau$. Thus, we have
    \begin{equation}
    \label{parasgrad}
    \frac{d L}{d \theta_k} =  \int_{k\tau+\tau}^{k\tau} -\mathbf{a}(t)^{\top} \frac{\partial f}{\partial \theta_k} d t,
    \end{equation}
    where the dynamics of the adjoint can be specified as:
    \begin{equation}
    \label{adjoint}
    \left\{
    \begin{aligned}
        \frac{d \mathbf{a}(t)}{d t} 
                    & = -\mathbf{a}(t)^{\top} \frac{\partial f}{\partial \mathbf{z}(t)}, ~t\in[k\tau, k\tau+\tau]\\
                    \mathbf{a}(l\tau)& =\mathbf{a}(l\tau) + \int_{k\tau+\tau}^{k\tau} -\mathbf{a}(t)^{\top} \frac{\partial f}{\partial \mathbf{z}(l\tau)}dt, \\ 
                    & ~l=0,1,\cdots,k,\\
    \end{aligned}
    \right.
    \end{equation}
where the backward initial condition $\mathbf{a}(T) = \frac{\partial L(\mathbf{h}(T))}{\partial \mathbf{z}(T)}$ and $k=n-1,n-2,\cdots,0$. 
\end{thm}

 We note that in \eqref{adjoint}, due to the skip connections, analogous to DenseNets \citep{huang2017densely},  the gradients are accumulated from multiple paths through the reversed skip connections in the backward direction, which likely renders the parameters optimized sufficiently.  Additionally, if the loss function $L(\mathbf{z}(T))$ depends on the states at different time points, viz., the new loss function $L(\mathbf{z}(t_0), \mathbf{z}(t_1),..., \mathbf{z}(t_N))$, we need to update instantly the adjoint state in the backward direction by adding the partial derivative of the loss at the observational time point, viz. $\mathbf{a}(t_i) =\mathbf{a}(t_i)+\frac{\partial L}{\partial \mathbf{z}(t_i)}$. For the specific tasks of classification and regression, refer to the section of \textbf{Experiments}. 

\section{Major Properties of NPCDDEs}
\label{sec4}
The NPCDDEs in \eqref{eqDNDDEgenral} and the UNPCDDEs in \eqref{eqDNDDEgenralunshared} generalize the ResNets and the NODEs as well. Also, they have strong connections with the Augmented NODEs. Moreover, the discontinuous nature of the NPCDDEs enables us to model complex dynamics beyond the NODEs, the Augmented NODEs, and the NDDEs. Lastly, the NPCDDEs are shown to enjoy advantages in 
computation over the NDDEs. In the sequel, we discuss these properties.  

\noindent
\paragraph{\fcircle[fill=black]{3pt} Both the ResNets and the NODEs are special cases of the UNPCDDEs in \eqref{eqDNDDEgenralunshared}.}
    We emphasize that any dimension-preserving neural networks (multi-layer residual blocks) are special cases of the UNPCDDEs. Actually, one can enforce the $\mathbf{z}(t), \mathbf{z}(\left\lfloor \frac{t-\tau}{\tau} \right\rfloor \tau), \mathbf{z}(\left\lfloor \frac{t-2\tau}{\tau} \right\rfloor \tau), \cdots, \mathbf{z}(0)$ as the dummy variables in the vector field of \eqref{eqDNDDEgenral} by assigning the weights connected to these variables to be zero, except for the variable $\mathbf{z}(\left\lfloor \frac{t}{\tau} \right\rfloor \tau)$. 
    Moreover, letting $\tau=1$ results in very simple unshared NPCDDEs as:
    \begin{equation}
    \label{eqDNDDEgenralsimple}
            \frac{d\mathbf{z}(t) }{d t} =  f(\mathbf{z}(k), \theta_k), ~t\in[k, k+1],~\mathbf{z}(0)=\mathbf{x}. 
    \end{equation}
    Due to the vector field of \eqref{eqDNDDEgenralsimple} keeping constant in each interval $[k, k+1]$, we have 
    \begin{equation}
    \label{eqDNDDEgenralsimple_res}
            \mathbf{z}(k+1) = \mathbf{z}(k) + f(\mathbf{z}(k), \theta_k),~ 
            \mathbf{z}(0)=\mathbf{x}, 
    \end{equation}
    which is exactly the form of the ResNets \eqref{eq1}. In addition, if we let $\mathbf{z}(\left\lfloor \frac{t}{\tau} \right\rfloor \tau), \mathbf{z}(\left\lfloor \frac{t-\tau}{\tau} \right\rfloor \tau), ..., \mathbf{z}(0)$ as the dummy variables in the vector field of \eqref{eqDNDDEgenral} and set $\theta_k\equiv \theta$, the UNPCDDEs in \eqref{eqDNDDEgenralunshared} indeed become the typical NODEs. Interestingly, though the NODEs are inspired by the ResNets, they are not equivalent to each other because of the limited modeling capability of the NODEs. But UNPCDDEs in \eqref{eqDNDDEgenralunshared} provides a more general form of the two. 
    
\noindent
\paragraph{\fcircle[fill=black]{3pt} Connection to Augmented NODEs}
The NPCDDEs in \eqref{eqDNDDEgenral} can be viewed as a particular form of the Augmented NODEs: 
\begin{equation}
\label{eqDNDDEgenralaug}
\left\{
    \begin{aligned}
        \frac{d\mathbf{z}(t) }{d t} &=  f(\mathbf{z}(t), \mathbf{z}_0(t), \mathbf{z}_1(t), ..., \mathbf{z}_n(t), \theta), t\in[0, T],\\
         \frac{d\mathbf{z}_0(t)}{d t} &=  \mathbf{0}, \mathbf{z}_0(t) =\mathbf{z}(\left\lfloor \frac{t}{\tau} \right\rfloor \tau), \\
         &\cdots\\
         \frac{d\mathbf{z}_n(t) }{d t} &=  \mathbf{0}, \mathbf{z}_n(t) =\mathbf{z}(\left\lfloor \frac{t-n\tau}{\tau} \right\rfloor \tau), \\
        \mathbf{z}(-n\tau)&=\cdots\mathbf{z}(-\tau)=\mathbf{z}(0)=\mathbf{x}. 
    \end{aligned}
\right.
\end{equation} 
Hence, we can apply the framework of the NODEs to coping with the NPCDDEs by solving the Augmented NODEs in \eqref{eqDNDDEgenralaug}.  It is worthwhile to emphasize that the Augmented NODEs in \eqref{eqDNDDEgenralaug} are not trivially equivalent to the traditional Augmented NODEs developed in \citep{dupont2019augmented}. In fact, the dynamics of $\mathbf{z}_i(t)$ in \eqref{eqDNDDEgenralaug} are piecewise-constant (but $\mathbf{z}(t)$ is continuous) and thus \textit{discontinuous} at each time instant $k\tau$, while the traditional Augmented NODEs still belong to the framework of NODEs whose dynamics are continuously evolving. The benefits of discontinuity are specified in the following.

\noindent
\paragraph{\fcircle[fill=black]{3pt} Discontinuity of the piecewise-constant delay(s)}
   {
    Notice that $\lfloor \cdot \rfloor$ used in the piecewise-constant delay(s) is a discontinuous function, which makes the first-order derivative of the function discontinuous at each key time point (i.e., integer multiple of the time delay).  This characteristic overcomes a huge limitation, the homeomorphisms/continuity of the trajectories produced by the NODEs, and thus enhances the flexibility of the NPCDDEs to handling plenty of complex dynamics (e.g., jumping derivatives and chaos evolving in the lower dimensional space). We will validate this advantage in the section of \textbf{Experiments}. 
    Additionally, the simple Euler scheme for the ODEs in \eqref{eq1} is actually a special PCDDE: $\frac{d\mathbf{z}(t) }{d t} = f(\mathbf{z}(\lfloor \frac{t}{\Delta t}\rfloor \Delta t))$ \citep{cooke1991survey}.   Based on the discontinuous nature, the approximation of the DDEs using the PCDDEs has been validated in \citep{cooke1991survey}.  Finally, such kind of discontinuous settings could be seen as typical forms of those discontinuous control strategies that are frequently used in control problems \citep{evans1983introduction, lewis2012optimal}. Actually, discontinuous control strategies can bring benefits on time and energy consumption \citep{sun2017closed}.
   }

{
\noindent
\paragraph{\fcircle[fill=black]{3pt} Computation advantages of NPCDDEs over NDDEs}
For solving the conventional NDDEs in \eqref{eqNDDEold}, we need to recompute the delay states in time using appropriate ODE solver \citep{zhu2021neural}, which requires $\mathcal{O}(n)$ memory and $\mathcal{O}(nK)$ computation, where $K$ is the adaptive depth of the ODE solver. On the contrary, for NPCDDEs in \eqref{eqDNDDEgenral} and the unshared NPCDDEs in \eqref{eqDNDDEgenral}, the delays are constant, and thus recomputing is not needed. As a result, for NPCDDEs (or UNPCDDEs), computational cost is approximately in orders of $\mathcal{O}(n)$ and $\mathcal{O}(K)$. Thus, the computational cost of NPCDDEs is cheaper than NDDEs.
}
 
     \begin{figure*}[htb]
      \vskip 0.0in
        \begin{center}
            \includegraphics[width=15cm]{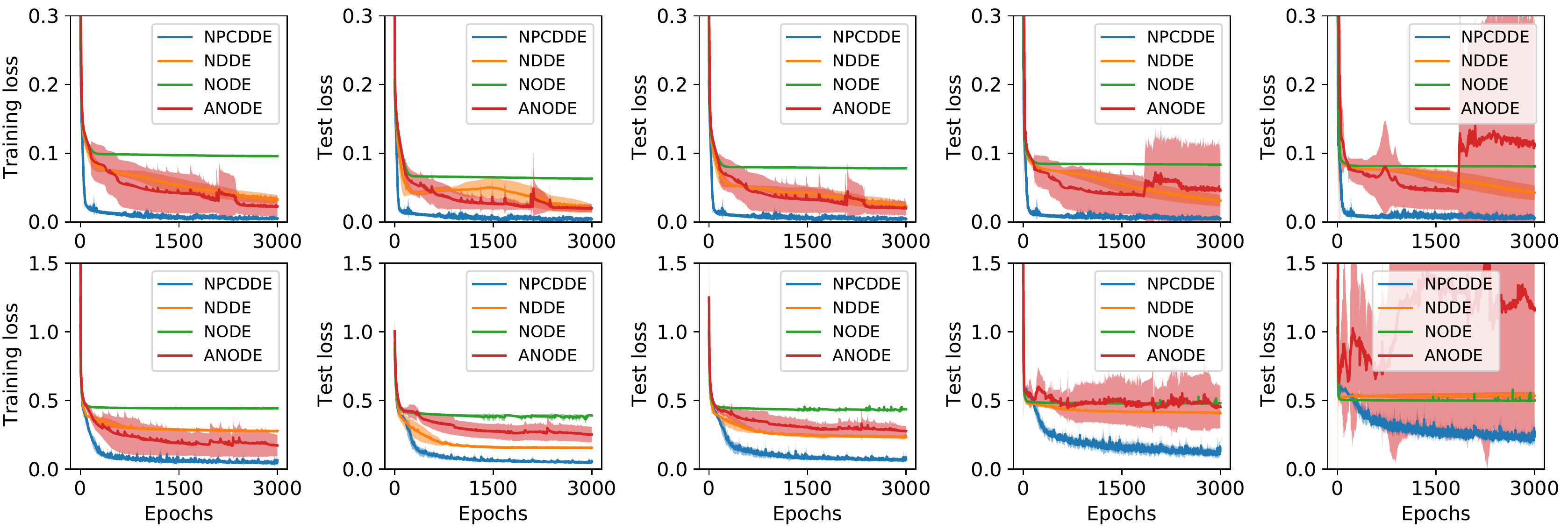}
        \end{center}
        \caption{
        {The training losses and the test losses of the piecewise-constant delay population dynamics \eqref{PCDDE1d} with the growth parameter $a=2.0$ (the oscillation regime, top) and $a=3.2$ (the chaos regime, bottom), respectively, by using the NPCDDEs, the NDDEs, the NODEs, and the Agumented NODEs (where the augmented dimension equals to $1$). The panels in the first column depict the training losses. The panels from the second column to the fifth column depict the test losses over the time intervals, respectively, with the lengths $1$, $2$, $5$, and $10$.}
        }
        \label{P_1d_fig}
        \vskip -0.1in
    \end{figure*}
    
\section{Experiments}   
\label{sec5}
\subsection{Population Dynamics: One-Dimensional PCDDE} \label{sec:toy_exp}
    We consider a 1-d PCDDE, which reads:  
    \begin{equation}
        \label{PCDDE1d}
        \frac{dx(t)}{dt} = a x(t) (1-x(\lfloor t \rfloor)), ~x(0)=x_0\geq 0.
    \end{equation}
    where the growth parameter $a>0$ \citep{1988A, cooke1991survey}. The above PCDDE~\eqref{PCDDE1d} is analogous to the well-known, first-order nonlinear logistic differential equation of one-dimension, which describes the growth dynamics of a single population and can be written as:
    \begin{equation}
        \label{logeq}
        \frac{dx(t)}{dt} = a x(t) (1-x(t)), ~x(0)=x_0\geq 0.
    \end{equation}
    Clearly, replacing the term $1-x(t)$ in the vector field of \eqref{logeq} by the term $1-x(\lfloor t \rfloor)$ results in the vector field of \eqref{PCDDE1d}. For each given $a>0$ and $x_0\geq 0$, if we consider the state $x(t)$ at the integer time instants $t=0,1,2,\cdots$, the corresponding discrete sequence, $x(0), x(1), x(2),\cdots$, satisfy the following discrete dynamical system:  
    \begin{equation}
        \label{1dmapppp}
        x(t+1) = x(t) e^{a(1-x(t))}, ~t=0,1,2,\cdots.
    \end{equation}
    Thus, we study the function
    \begin{equation}
        \label{1dmap}
        f_a(x) = x e^{a(1-x)}, ~x\in[0,\infty). 
    \end{equation}
    Direct computation indicates that the function $f_a(\cdot)$ in \eqref{1dmap} is a $C^1$-unimodal map in $[0,\infty]$ with its maximal value as $f_a(x^*)=f_a(\frac{1}{a})$.  Thus, $[0, \frac{1}{a}]$ is a strictly increasing regime of this function while $[\frac{1}{a}, \infty)$ is a strictly decreasing regime.  As pointed out in \citep{1988A, cooke1991survey}, the discrete dynamical system \eqref{1dmapppp} can exhibit complex dynamics including chaos. More precisely, at $a^*=3.11670...$, the solution of \eqref{1dmapppp} with the initial value $x(0)=x_0=\frac{1}{a^*}$ is periodic and asymptotically stable with a period of three, so that $f_{a^*}\circ f_{a^*} \circ f_{a^*} (x_0)=x_0$.  This further 
    implies that the map with the adjustable parameter $a$ admits period-doubling bifurcations and thus has chaotic dynamics according to the well-known Sharkovskii Theorem \citep{li1975period, 1988A, cooke1991survey}. Moreover, since the discrete dynamical system \eqref{1dmapppp} could be regarded as the sampled system with integer sampling time instants from the original PCDDE \eqref{PCDDE1d}, this PCDDE exhibits chaotic as well for $a$ in the vicinity of $a^*$. We thereby test the NODEs, the NDDEs, the NPCDDEs, and the Augmented NODEs on the piecewise-constant delay population dynamics \eqref{PCDDE1d}, respectively, with $a=2.0$ and $a=3.2$, which corresponds to two regimes of oscillation and chaos.  
    Moreover, as can be seen from Fig. \ref{P_1d_fig}, the training losses and the test losses of the NPCDDEs decrease significantly, compared to those of the other models. Additionally, in the oscillation regime, the losses of the NPCDDEs approach a very low level in both training and test stages, while in the chaos regime, the NPCDDEs can achieve short-term prediction in an accurate manner.  Naturally, it is hard to achieve long-term prediction because of the sensitive independence of initial conditions in a chaotic system.  Here, for training, we produce $100$ time series from different initial states in the time interval $[0, 3]$ with $0.1$ as the sampling period.  Thus, still with $0.1$ as the sampling period, we use the final states of the training data as the initial states for the $100$ test time series in the next time interval $[3, 13]$.  More specific configurations for our numerical experiments are provided in the supplementary material.

\subsection{Image datasets}
\label{secimage}
    We conduct experiments on several image datasets, including MNIST, CIFAR10, SVHN, by using the (unshared) NPCDDEs and the other baselines.  In the experiments, we follow the setup in the work \citep{zhu2021neural}.  For a fair comparison, we construct all models without augmenting the input space, and for the NDDEs, we assume that the initial function keeps constant (i.e., the initial function $\phi(t)=\mbox{input}$ for $t\leq 0$), which is different from the initial function used for the NDDEs in \citep{zhu2021neural}. We note that our models are orthogonal to these models, since one can also augment the
    input space and model the initial state as the feature of an NODE in the framework of NPCDDEs.  Additionally, the number of the parameters for all models are almost the same ($84$k params for MNIST, $107$k params for CIFAR10 and SVHN).  Notably, the vector fields of all the models are parameterized with the convolutional architectures \citep{dupont2019augmented, zhu2021neural}, where the arguments that appeared in the vector fields are concatenated and then fed into the convolutional neural networks (CNNs).  For example, for the NDDEs, the vector field is $f(\mbox{concat}(\mathbf{z}(t), \mathbf{z}(t-\tau)), \theta)$, where $\mbox{concat}(\cdot,\cdot)$ is a concatenation operator for two tensors on the channel dimension. Moreover, the initial states for these models are just the images from the datasets.  It is observed that our models outperform the baselines on these datasets. The detailed test accuracies are shown, respectively, in Tab.~\ref{table}. For the specific training configurations for all the models and more experiments equipped with augmentation \cite{dupont2019augmented}, please refer to the supplementary material.

\begin{table}[t]
    \vskip -0.1in
        \caption{The test accuracies with their standard deviations over 5 realizations of different models on the image datasets.  In the first column, the integer $i$ in NPCDDE$i$ or UNPCDDE$i$ means that $n=i$ for the NPCDDEs in \eqref{eqDNDDEgenral} or for the UNPCDDEs in \eqref{eqDNDDEgenralunshared}. The results for the NODEs and NDDEs are reported in \citep{zhu2021neural}. The final time $T$ for all models is assigned as $1$.
        }
    \begin{center}
     \resizebox{\linewidth}{!}{
    \begin{tabular}{llll}
    \hline\hline
    \multicolumn{1}{c}{\bf ~}  &\multicolumn{1}{c}{CIFAR10} &\multicolumn{1}{c}{MNIST} &\multicolumn{1}{c}{SVHN}
    \\ \hline 
    NODE    &$53.92\%\pm0.67$   &$96.21\%\pm0.66$   &$80.66\%\pm0.56$\\
    NDDE    &$55.69\%\pm0.39$   &$96.22\%\pm0.55$   &$81.49\%\pm0.09$\\
    NPCDDE2 (ours)  &$56.03\%\pm0.25$   &$97.32\%\pm0.30$   &$82.63\%\pm0.36$\\
    UNPCDDE2 (ours)   &$56.22\%\pm0.42$   &$97.43\%\pm0.13$   &$82.99\%\pm0.23$\\
    NPCDDE3 (ours)   &$56.34\%\pm0.51$   &$97.34\%\pm0.10$   &$82.38\%\pm0.35$\\
    UNPCDDE3 (ours)   &$56.09\%\pm0.37$   &$97.52\%\pm0.14$   &$83.19\%\pm0.32$\\
    NPCDDE5 (ours)   &$56.59\%\pm0.44$   &$97.40\%\pm0.19$   &$82.62\%\pm0.69$\\
    UNPCDDE5 (ours)   &${\bf56.73\%\pm0.54}$   &${\bf 97.69\%\pm0.13}$   &${\bf 83.45\%\pm0.38}$\\
    \hline\hline
    \end{tabular}
    }
    \end{center}
    \label{table}
    \vskip -0.2in
\end{table}

\section{Discussion}
As shown above, the NPCDDEs achieve good performances not only on the 1-d PCDDE example but on the image datasets as well.  However, such NPCDDEs are not the perfect framework, still having some limitations.  Here, we suggest several directions for future study, including: 1) For an NPCDDE, seeking a good strategy to determine the number of the skip connections and the specific value of each delay for different tasks, 2) applying the NPCDDEs to the other suitable real-world datasets, such as the time series with the piecewise-constant delay effects, 3) providing more analytical results for the NPCDDEs to guarantee the stability and robustness, 
and 4) leveraging the optimal control theory \cite{pontryagin1962mathematical} for dynamical systems to further promote the performance of neural networks.

\section{Conclusion}
In this article, we have articulated a framework of the NPCDDEs, which is mainly inspired by several previous frameworks, including the NODEs, the NDDEs,  
and the PCDDEs. The NPCDDEs own not only the provable capability of universal approximation but also the outstanding power of nonlinear representations.   Also, we have derived the backward gradients along with the adjoint dynamics for the NPCDDEs.  We have emphasized that both the ResNets and the NODEs are the special cases of the NPCDDEs and that the NPCDDEs are of a more general framework compared to the existing models.  Finally, we have demonstrated that the NPCDDEs outperform the several existing frameworks on  representative image datasets (MNIST, CIFAR10, and SVHN).  All these 
suggest that integrating the elements of dynamical systems with different kinds of neural networks is indeed beneficial to creating and promoting the frameworks of deep learning using continuous-depth structures.

\section{Acknowledgments}
We thank the anonymous reviewers for their valuable and constructive comments that helped us to improve the work. Q.Z is supported by the STCSM (No. 21511100200). 
W.L. is supported by the National Key R\&D Program of China (No. 2018YFC0116600), by the National Natural Science Foundation of China (Nos. 11925103 and 61773125), and by the STCSM (Nos. 19511132000, 19511101404, and 2021SHZDZX0103).




\bibliography{aaai22}

\clearpage
\appendix

\section{Proof of Proposition~\ref{eq:x^2}}
The proof has been provided in \citep{Radovanovic2007FunctionalE}. However, for convenience and compactness of the current work, we restate the proof here.  First, we rewrite the functional equation as
\begin{equation}
    f(f(x)) = x^2-2.
\end{equation}
Notice that the function $g$ of the right-hand side has exactly $2$ fixed points and that the function $g \circ g$
has exactly $4$ fixed points. Now we are to prove that there is no function $f$ such that $f \circ f = g$ by contradiction. To this end, let $a$, $b$ be the fixed points of $g$, and $a$, $b$, $c$, $d$ be the fixed points of $g \circ g$, and
assume that
$g(c) = y$. Then, $c = g(g(c)) = g(y)$, so that $g(g(y)) = g(c) = y$ and $y$ has to be on of the fixed points of
$g \circ g$. If $y = a$, then, from $a = g(a) = g(y) = c$, we get a contradiction. Similarly, we can get $y \neq b$, and, due to $y \neq c$,
we get $y = d$. Thus, $g(c) = d$ and $g(d) = c$. Furthermore, we have $g(f(x)) = f(f(f(x))) = f(g(x))$.
Let $x_0 \in \{a,b\}$. We immediately have $f(x_0) = f(g(x_0)) = g(f(x_0))$. Hence, $f(x_0) \in \{a,b\}$. Similarly,
if $x_1 \in \{a,b,c,d\}$, we get $f(x_1) \in \{a,b,c,d\}$. This is impossible, which we are in a position to validate. Taking
$f(c) = a$ implies $f(a) = f(f(c)) = g(c) = d$.  This is clearly impossible. Similarly, we have $f(c) \neq b$ and
$f(c) \neq c$ (otherwise, $g(c)= c$).  Hence, $f(c) = d$, which implies $f(d) = f(f(c)) = g(c) = d$.
This, however, becomes a contradiction again, proving that the required $f$ does not exist.

\section{Proof of Theorem~\ref{thm1}}
    Here, the proof is provided in a constructive manner. Assume that, for any map $\mathbf{x} \mapsto F(\mathbf{x})$, there exists a neural network (i.e., one hidden layer feed-forward neural network), such that its input is the concatenation of $\mathbf{z}(\left\lfloor \frac{t}{\tau} \right\rfloor \tau)$ and $\mathbf{z}(\left\lfloor \frac{t-\tau}{\tau} \right\rfloor \tau)$. In addition, we assign all the weights directly connecting to $\mathbf{z}(\left\lfloor \frac{t}{\tau} \right\rfloor \tau)$ as zero.  As such, the $\mathbf{z}(\left\lfloor \frac{t}{\tau} \right\rfloor \tau)$ becomes a dummy variable, resulting in the intrinsic vector field in the  NPCDDEs $(\ref{eqDNDDE})$ is in the form of $f(\mathbf{z}(\left\lfloor \frac{t-\tau}{\tau} \right\rfloor \tau), \theta)$. Hence, we only require the neural network, $f$, to approximate the function $\frac{F(\mathbf{x})-\mathbf{x}}{2\tau}$. Finally, using the assumption $\mathbf{z}(-\tau)=\mathbf{z}(0)=\mathbf{x}$, we have $\mathbf{z}(\tau)=\mathbf{z}(0) + \tau \cdot f(\mathbf{z}(-\tau), \theta) \approx x + \tau \cdot \frac{F(\mathbf{x})-\mathbf{x}}{2\tau} = \frac{F(\mathbf{x})+\mathbf{x}}{2}$ and then $\mathbf{z}(2\tau)=\mathbf{z}(\tau) + \tau \cdot f(\mathbf{z}(0), \theta)\approx \frac{F(\mathbf{x})+\mathbf{x}}{2} + \tau \cdot \frac{F(\mathbf{x})-\mathbf{x}}{2\tau}=F(\mathbf{x})$. Therefore, the proof is completed.
    
\section{Proof of Proposition~\ref{prop2}}
    The proof is analogous to that of Theorem~\ref{thm1}.  We assume the variables $\mathbf{z}(\left\lfloor \frac{t}{\tau} \right\rfloor \tau), \mathbf{z}(\left\lfloor \frac{t-\tau}{\tau} \right\rfloor \tau), ..., \mathbf{z}(\left\lfloor \frac{t-n\tau+\tau}{\tau} \right\rfloor \tau)$ as the dummy variables, except for $\mathbf{z}(\left\lfloor \frac{t-n\tau}{\tau} \right\rfloor \tau)$.  Thus, the intrinsic vector field in the  NPCDDEs $(\ref{eqDNDDE})$ can be rewritten as $f(\mathbf{z}(\left\lfloor \frac{t-n\tau}{\tau} \right\rfloor \tau), \theta)$. Hence, we only require the neural network, $f$, to approximate the function  $\frac{F(\mathbf{x})-\mathbf{x}}{n\tau}$. Finally, using the assumption $\mathbf{z}(-n\tau)=\cdots=\mathbf{z}(-\tau)=\mathbf{z}(0)=\mathbf{x}$, we obtain $\mathbf{z}(\tau)=\mathbf{z}(0) + \tau \cdot f(\mathbf{z}(-n\tau), \theta) \approx x + \tau \cdot \frac{F(\mathbf{x})-\mathbf{x}}{n\tau} = \frac{F(\mathbf{x})+(n-1)\mathbf{x}}{n}$. Thus, $\mathbf{z}(2\tau)=\mathbf{z}(\tau) + \tau \cdot f(\mathbf{z}(-n\tau+\tau), \theta)\approx \frac{F(\mathbf{x})+(n-1)\mathbf{x}}{n} + \tau \cdot \frac{F(\mathbf{x})-\mathbf{x}}{n\tau}=\frac{2F(\mathbf{x})+(n-2)\mathbf{x}}{n}$. Consequently, we have $\mathbf{z}(n\tau)=F(\mathbf{x})$, which completes the proof.

\section{Proof of Theorem~\ref{thm2}} 
    \subsection{Proof 1}
    The main mission of this proof is to obtain the adjoint dynamics of the UNPCDDEs in \eqref{eqDNDDEgenralunshared} by using the limitation of the traditional backpropagation. We first employ the Euler discretization with sufficiently small step size $\epsilon$ in the interval $[n\tau-\tau, n\tau]$ to get the discrete dynamics of the UNPCDDEs in \eqref{eqDNDDEgenralunshared}.  Such a discrete form can be written as:
    \begin{equation}
        \begin{aligned}
            \mathbf{z}(t+\epsilon) =& \mathbf{z}(t) + \epsilon \cdot f(\mathbf{z}(t), \mathbf{z}(\left\lfloor \frac{t}{\tau} \right\rfloor \tau), \mathbf{z}(\left\lfloor \frac{t-\tau}{\tau} \right\rfloor \tau), \cdots, \\
            &\mathbf{z}(0), \theta_{n-1}), ~t\in[n\tau-\tau, n\tau].\\
        \end{aligned}
    \end{equation}
    According to the definition of the adjoint $\mathbf{a}(t):=\frac{\partial L}{\partial \mathbf{z}(t)}$ and applying the chain rule of derivatives, we obtain 
    \begin{equation}
        \begin{aligned}
            \mathbf{a}(t) &= \frac{\partial{L}}{\partial{\mathbf{z}(t+\epsilon)}}\frac{\partial{\mathbf{z}(t+ \epsilon)}}{\partial{\mathbf{z}(t)}}\\
            & = \mathbf{a}(t+\epsilon)\left(I+\epsilon\cdot\frac{\partial f}{\mathbf{z}(t)}\right).
        \end{aligned}
    \end{equation}
    This further yields:
    \begin{equation}
        \frac{d\mathbf{a}(t)}{dt} = \lim_{\epsilon \rightarrow 0}\frac{\mathbf{a}(t+\epsilon)-\mathbf{a}(t)}{\epsilon}=-\mathbf{a}(t)\frac{\partial f}{\mathbf{z}(t)},
    \end{equation}
    for $t\in[n\tau-\tau, n\tau]$, where the backward initial state $\mathbf{a}(n\tau) =\frac{\partial L}{\partial \mathbf{z}(n\tau)}$. 
    For the gradients with respect to the vector of parameters $\theta_{n-1}$, we obtain 
    \begin{equation}
        \begin{aligned}
            \frac{d L}{d\theta_{n-1}} &= \lim_{\epsilon \rightarrow 0} \sum \frac{\partial L}{\partial\mathbf{z}(t+\epsilon)} \frac{\partial\mathbf{z}(t+\epsilon)}{\partial\theta_{n-1}}\\
             &= \lim_{\epsilon \rightarrow 0} \sum \mathbf{a}(t+\epsilon) \frac{\partial f}{\partial\theta_{n-1}} \epsilon\\
             & = \int_{n\tau}^{n\tau-\tau} -\mathbf{a}(t)\frac{\partial f}{\partial \theta_{n-1}} dt.
        \end{aligned}
    \end{equation}
    Notice that the piecewise-constant delay(s) $\mathbf{z}(\left\lfloor \frac{t}{\tau} \right\rfloor \tau), \mathbf{z}(\left\lfloor \frac{t-\tau}{\tau} \right\rfloor \tau), \cdots, \mathbf{z}(0)$ (equivalently,  $\mathbf{z}(n\tau-\tau), \mathbf{z}(n\tau-2\tau), \cdots, \mathbf{z}(0)$) appear in the vector field.  We therefore need to add the gradients from the interval $[n\tau-\tau, n\tau]$, 
    \begin{equation}
        \begin{aligned}
            \frac{\partial L}{\partial \mathbf{z}(l\tau)} &= \frac{\partial L}{\partial z(l\tau)} + \lim_{\epsilon \rightarrow 0} \sum   \frac{\partial L}{\partial\mathbf{z}(t+\epsilon)} \frac{\partial\mathbf{z}(t+\epsilon)}{\partial \mathbf{z}(l\tau)},\\
             &= \frac{\partial L}{\partial z(l\tau)} + \lim_{\epsilon \rightarrow 0} \sum \mathbf{a}(t+\epsilon) \frac{\partial f}{\partial \mathbf{z}(l\tau)}\epsilon,\\
            &= \frac{\partial L}{\partial z(l\tau)} + \int_{n\tau}^{n\tau-\tau} -\mathbf{a}(t)^{\top} \frac{\partial f}{\partial \mathbf{z}(l\tau)}dt,
        \end{aligned}
    \end{equation}
    where $l=0,1,\cdots, n-1$.
    
    Analogously, we successively obtain the backward gradients of interest for the intervals $[n\tau-2\tau, n\tau -\tau],\cdots, [\tau, 2\tau], [0,\tau]$. To summarize, we get 
    \begin{equation}
    \label{sup_parasgrad}
    \frac{d L}{d \theta_k} =  \int_{k\tau+\tau}^{k\tau} -\mathbf{a}(t)^{\top} \frac{\partial f}{\partial \theta_k} d t,
    \end{equation}
    where the dynamics of the adjoint can be written as:
    \begin{equation}
    \label{sup_adj}
    \left\{
    \begin{aligned}
        \frac{d \mathbf{a}(t)}{d t} 
                    & = -\mathbf{a}(t)^{\top} \frac{\partial f}{\partial \mathbf{z}(t)}, ~t\in[k\tau, k\tau+\tau]\\
                    \mathbf{a}(l\tau)& =\mathbf{a}(l\tau) + \int_{k\tau+\tau}^{k\tau} -\mathbf{a}(t)^{\top} \frac{\partial f}{\partial \mathbf{z}(l\tau)}dt, \\ 
                    & ~l=0,1,\cdots,k,\\
    \end{aligned}
    \right.
    \end{equation}
    where the backward initial condition $\mathbf{a}(T) = \frac{\partial L(\mathbf{h}(T))}{\partial \mathbf{z}(T)}$ and $k=n-1,n-2,\cdots,0$. The proof is consequently completed.
    
    \subsection{Proof 2}
    Actually, the UNPCDDEs in \eqref{eqDNDDEgenralunshared} at each interval $[k\tau, k\tau+\tau]$ with $k=n-1,n-2,...,0$ are equivalent to the following Augmented NODEs:
    \begin{equation}
    \label{eqDNDDEgenralaug_unshared}
    \left\{
    \begin{aligned}
        \frac{d\mathbf{z}(t) }{d t} &=  f(\mathbf{z}(t), \mathbf{z}_k(t), \mathbf{z}_{k-1}(t), ..., \mathbf{z}_0(t), \theta_k), \\
         \frac{d\mathbf{z}_k(t)}{d t} &=  \mathbf{0}, ~\mathbf{z}_k(k\tau) =\mathbf{z}(k\tau), \\
         &\cdots\\
         \frac{d\mathbf{z}_0(t) }{d t} &=  \mathbf{0}, ~\mathbf{z}_0(t) =\mathbf{z}(0), \\
    \end{aligned}
    \right.
    \end{equation}
    or
    \begin{equation}\label{aguee}
         \frac{d {\mathbf{z}}_{\rm aug}(t)}{d t} =f_{\rm aug}({\mathbf{z}}_{\rm aug}(t),\theta_k).
    \end{equation}
    Thereby, it is direct to employ the framework of the NODEs \citep{chen2018neural}
    to solve \eqref{aguee}. Then, we have the gradients with respect to the parameters as:
     \begin{equation}
    \label{sup_parasgradaug}
    \frac{d L}{d \theta_k} =  \int_{k\tau+\tau}^{k\tau} -\mathbf{a}_{\rm aug}(t)^{\top} \frac{\partial f_{\rm aug}}{\partial \theta_k} d t,
    \end{equation}
    where the augmented adjoint dynamics obey:  
     \begin{equation}
        \label{sup_adj_aug}
         \frac{d {\mathbf{a}}_{\rm aug}(t)}{d t} =-{\mathbf{a}}_{\rm aug}(t) \frac{\partial f_{\rm aug}({\mathbf{z}}_{\rm aug}(t),\theta_k)}{\partial {\mathbf{z}}_{\rm aug}(t)}
    \end{equation}
    Now, it can be directly validated that Eqs.~\eqref{sup_parasgradaug} and \eqref{sup_adj_aug} are equivalent, respectively, to Eqs.~\eqref{sup_parasgrad} and \eqref{sup_adj}. The proof is therefore completed.

\section{Experimental details}
To make a fair comparison in our experiments, we use the setups almost similar to those used in \citep{dupont2019augmented,zhu2021neural}. 

\subsection{Piecewise-constant delay population dynamics}
    For the 1-d piecewise-constant delay population dynamics $\frac{dx(t)}{dt} = a x(t) (1-x(\lfloor t \rfloor))$ with $x(0)=x_0$, we choose the parameter $a=2.0$ (resp. $3.2$) corresponding to oscillation (resp. chaos) dynamics. 
    We model the system by applying the NODE, the ANODE, the NDDE, and the NPCDDE, whose structures are specified, respectively, as follows: 
    \begin{enumerate}
    \item Structure of the NODE:  The vector field is modeled as 
    \begin{equation*}
        f(x) = W_{\rm out}\tanh(W\tanh(W_{\rm in} x)),
    \end{equation*}
    where $W_{\rm in}\in \mathbb{R}^{10\times 1}$, $W\in \mathbb{R}^{10\times 10}$, and  $W_{\rm out}\in \mathbb{R}^{1\times 10}$;
    \item Structure of the ANODE: The vector field is modeled as
    \begin{equation*}
        f(\mathbf{x}_{\rm aug}(t)) = W_{\rm out} \tanh(W\tanh(W_{\rm in}\mathbf{x}_{\rm aug})),
    \end{equation*}
    where $W_{\rm in}\in \mathbb{R}^{10\times 2}$, $W\in \mathbb{R}^{10\times 10}$,  $W_{\rm out}\in \mathbb{R}^{2\times 10}$, and the augmented dimension is equal to $1$. Notably, we require to align the augmented trajectories with the target trajectories to be regressed. To achieve it, we choose the data in the first component and exclude the data in the augmented component (i.e., simply projecting the augmented data into the space of the target data). 
     \item Structure of the NDDE: The vector field is modeled as 
    \begin{equation*}
        \begin{aligned}
             &f(x(t), x(t-\tau)) \\
        = &W_{{\rm out}} \tanh(W\tanh(W_{\rm in} \mbox{\rm concat}(x(t), x(t-\tau)))),
        \end{aligned}
    \end{equation*}
    where $\tau=1$, $W_{\rm in}\in \mathbb{R}^{10\times 2}$, $W\in \mathbb{R}^{10\times 10}$, and  $W_{\rm out}\in \mathbb{R}^{1\times 10}$.
     \item Structure of the NPCDDE:  the vector field is modeled as 
    \begin{equation*}
        \begin{aligned}
             &f(x(t), x(\lfloor t \rfloor)) \\
        = &W_{\rm out} \tanh(W\tanh(W_{\rm in} \mbox{concat}(x(t), x(\lfloor t \rfloor)))),
        \end{aligned}
    \end{equation*}
    where $W_{\rm in}\in \mathbb{R}^{10\times 2}$, $W\in \mathbb{R}^{10\times 10}$, and  $W_{\rm out}\in \mathbb{R}^{1\times 10}$.
    \end{enumerate}
    We train each model for 5 runs with different random seeds. For each run, we set the hyperparameters as: $3000$, the iteration number, and $0.01$, the learning rate for the Adam optimizer.
    
    \begin{figure*}[htb]
        \begin{center}
            \centerline{\includegraphics[width=0.78\textwidth]{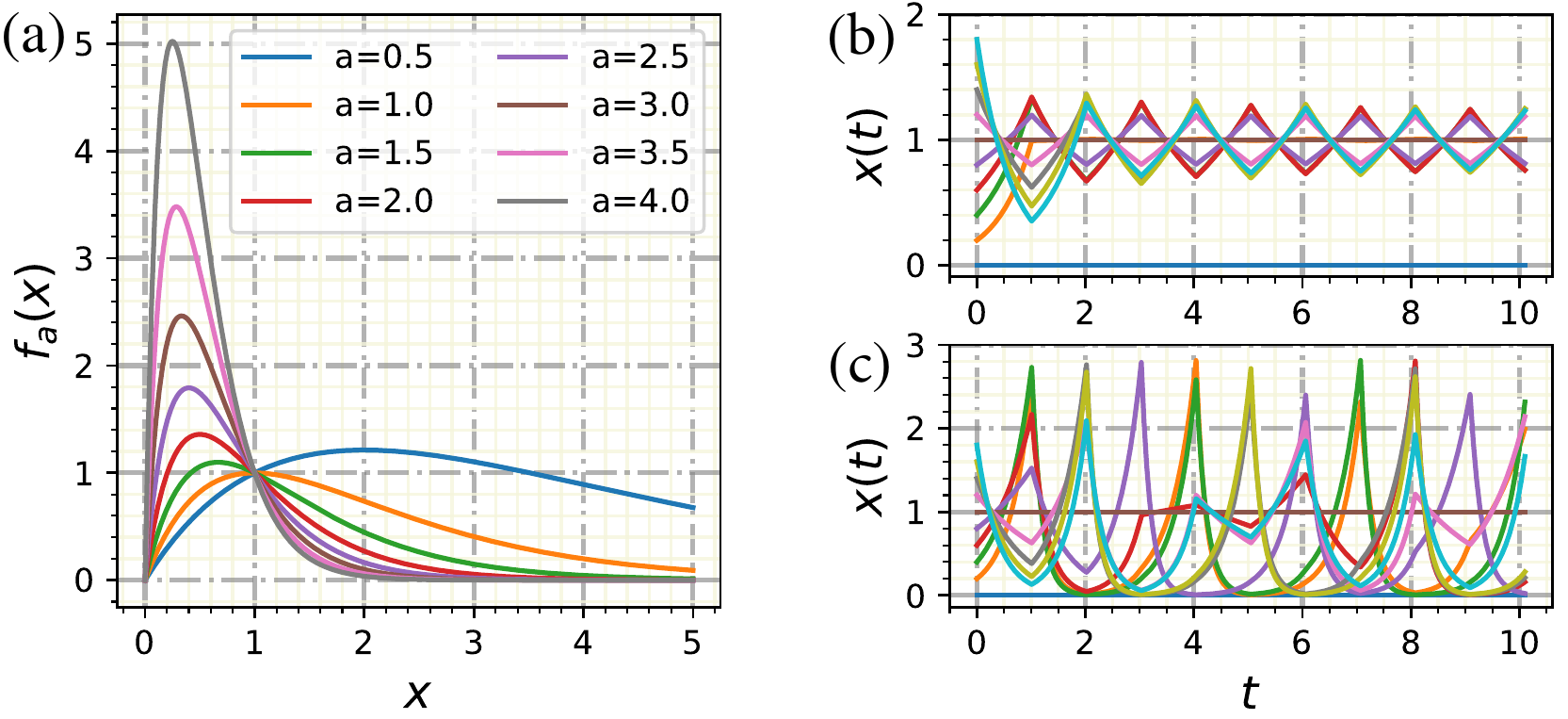}}
            \caption{(a) The curves of the function $f_a(x)$ defined in \eqref{1dmap} 
            with the growth parameter $a$ taking its value, respectively, from the set $\{0.5, 1.0, 1.5, 2.0, 2.5, 3.0, 3.5, 4.0\}$. The oscillation regime with $a=2.0$ (b) and the chaos regime  with $a=3.2$ (b), where the trajectories from $10$ different initial states
            are depicted for each regime.}
        \label{fig_fa_Osci_Chao}
        \end{center}
        \vskip -0.1in
    \end{figure*}
    
    \subsection{Image datasets}
    
    As for the image experiments using the NODE and the NDDE, the structures of our models are adapted from \citep{zhu2021neural}.  More precisely, we apply the convolution block with the following structures and dimensions:
    \begin{itemize}
        \item $1 \times 1$ conv, $k$ filters, $0$ paddings,
        \item ReLU activation function,
        \item $3 \times 3$ conv, $k$ filters, $1$ paddings,
        \item ReLU activation function,
        \item $1 \times 1$ conv, $c$ filters, $0$ paddings,
    \end{itemize}
    where $k$ is different for each model and $c$ is the channel number of the input images. In Tab.~\ref{table2}, we specify the information for each model. As can be seen, we fix approximately the same number of the parameters for each model. We select the hyperparamters for the Adam optimizer as: 1e-3, the learning rate, $256$, the batch size, and $30$, the number of the training epochs.

    \begin{table}[htb]
    \vskip -0.1in
        \caption{The test accuracies with their standard deviations over 5 realizations of different models with augmentation on the image datasets.  In the first column,  A$p$ with $p=1,2,4$ indicates the augmentation of the image space $\mathbb{R}^{c\times h \times w} \rightarrow \mathbb{R}^{ (c+p)\times h \times w}$, and the integer $j$ in NPCDDE$j$ or UNPCDDE$j$ means that $n=j$ for the NPCDDEs in \eqref{eqDNDDEgenral} or for the UNPCDDEs in \eqref{eqDNDDEgenralunshared}. The results for the Augmented NODEs and the Augmented NDDEs are reported in \citep{zhu2021neural}. The final time $T$ for all the models is assigned as $1$.
        }
    \begin{center}
     \resizebox{\linewidth}{!}{
    \begin{tabular}{llll}
    \hline\hline
    \multicolumn{1}{c}{\bf ~}  &\multicolumn{1}{c}{CIFAR10} &\multicolumn{1}{c}{MNIST} &\multicolumn{1}{c}{SVHN}
    \\ \hline 
    \hline
    A1+NODE   &$56.14\%\pm0.48$   &$97.89\%\pm0.14$   &$81.17\%\pm0.29$\\
    A1+NDDE   &$56.83\%\pm0.60$   &$97.83\%\pm0.07$   &$82.46\%\pm0.28$\\
    A1+NPCDDE5   &$57.05\%\pm 0.45$   &$97.98\%\pm 0.13$   &$83.19\%\pm 0.34$\\
    A1+UNPCDDE5   &${\bf58.00\%\pm 0.38}$   &${\bf98.10\%\pm 0.06}$   &${\bf83.62\%\pm 0.50}$\\
    \hline
    A2+NODE   &$57.27\%\pm0.46$   &${\bf98.25\%\pm0.08}$   &$81.73\%\pm0.92$\\
    A2+NDDE   &$58.13\%\pm0.32$   &$98.22\%\pm0.04$   &$82.43\%\pm0.26$\\
    A2+NPCDDE5   &$58.53\%\pm 0.40$   &$98.20\%\pm 0.10$   &$82.98\%\pm 0.37$\\
    A2+UNPCDDE5   &${\bf59.11\%\pm 0.49}$   &$98.23\%\pm 0.05$   &${\bf83.51\%\pm 0.49}$\\
    \hline
    A4+NODE   &$58.93\%\pm0.33$   &$98.33\%\pm0.12$   &$82.72\%\pm0.6$\\
    A4+NDDE   &$59.35\%\pm0.48$   &$98.31\%\pm0.03$   &$82.87\%\pm0.55$\\
    A4+NPCDDE5   &$59.71\%\pm 0.49$   &${\bf98.35\%\pm 0.10}$   &$83.57\%\pm 0.50$\\
    A4+UNPCDDE5   &${\bf60.20\%\pm 0.77}$   &$98.27\%\pm 0.08$   &${\bf84.16\%\pm 0.32}$\\
    \hline\hline
    \end{tabular}
    }
    \end{center}
    \label{acc_more}
    \vskip -0.2in
    \end{table}

    \begin{table}[htb]
    \caption{The number of the filters and the whole parameters in each model used for CIFAR10, MNIST, and SVHN. 
        }
        \begin{center}
        \resizebox{\linewidth}{!}{
        \begin{tabular}{llll}
        \hline\hline
        \multicolumn{1}{c}{\bf ~}  &\multicolumn{1}{c}{CIFAR10} &\multicolumn{1}{c}{MNIST} &\multicolumn{1}{c}{SVHN}
        \\ \hline 
        NODE   &$92/107645$   &$92/84395$   &$92/107645$\\
        NDDE   &$92/107921$   &$92/84487$   &$92/107921$\\
        NPCDDE2   &$91/106536$   &$91/82926$   &$91/106536$\\
        UNPCDDE2   &$64/106064$   &$64/82284$   &$64/106064$\\
        NPCDDE3   &$91/106809$   &$91/83017$   &$91/106809$\\
        UNPCDDE3   &$52/105931$   &$52/81797$   &$52/105931$\\
        NPCDDE5   &$91/107355$   &$91/83199$   &$91/107355$\\
        UNPCDDE5   &$40/106145$   &$40/81255$   &$40/106145$\\
        \hline
        A1+NODE   &$86/108398$   &$87/84335$   &$86/108398$\\
        A1+NDDE   &$85/107189$   &$86/82944$   &$85/107189$\\
        A1+NPCDDE5   &$84/106998$   &$86/83632$   &$84/106998$\\
        A1+UNPCDDE5   &$37/106665$   &$38/82960$   &$37/106665$\\
        \hline
        A2+NODE   &$79/108332$   &$81/83230$   &$79/108332$\\
        A2+NDDE   &$78/107297$   &$81/83473$   &$78/107297$\\
        A2+NPCDDE5   &$77/107425$   &$80/82973$   &$77/107425$\\
        A2+UNPCDDE5   &$34/107845$   &$35/81645$   &$34/107845$\\
        \hline
        A4+NODE   &$63/108426$   &$70/84155$   &$63/108426$\\
        A4+NDDE   &$62/107719$   &$69/83237$   &$62/107719$\\
        A4+NPCDDE5   &$61/108297$   &$68/83347$   &$61/108297$\\
        A4+UNPCDDE5   &$26/106955$   &$30/83785$   &$26/106955$\\
        \hline\hline
        \end{tabular}
        }
        \end{center}
        \label{table2}
    \end{table}
    
    \begin{figure*}[htb]
        \begin{center}
            \includegraphics[width=17cm]{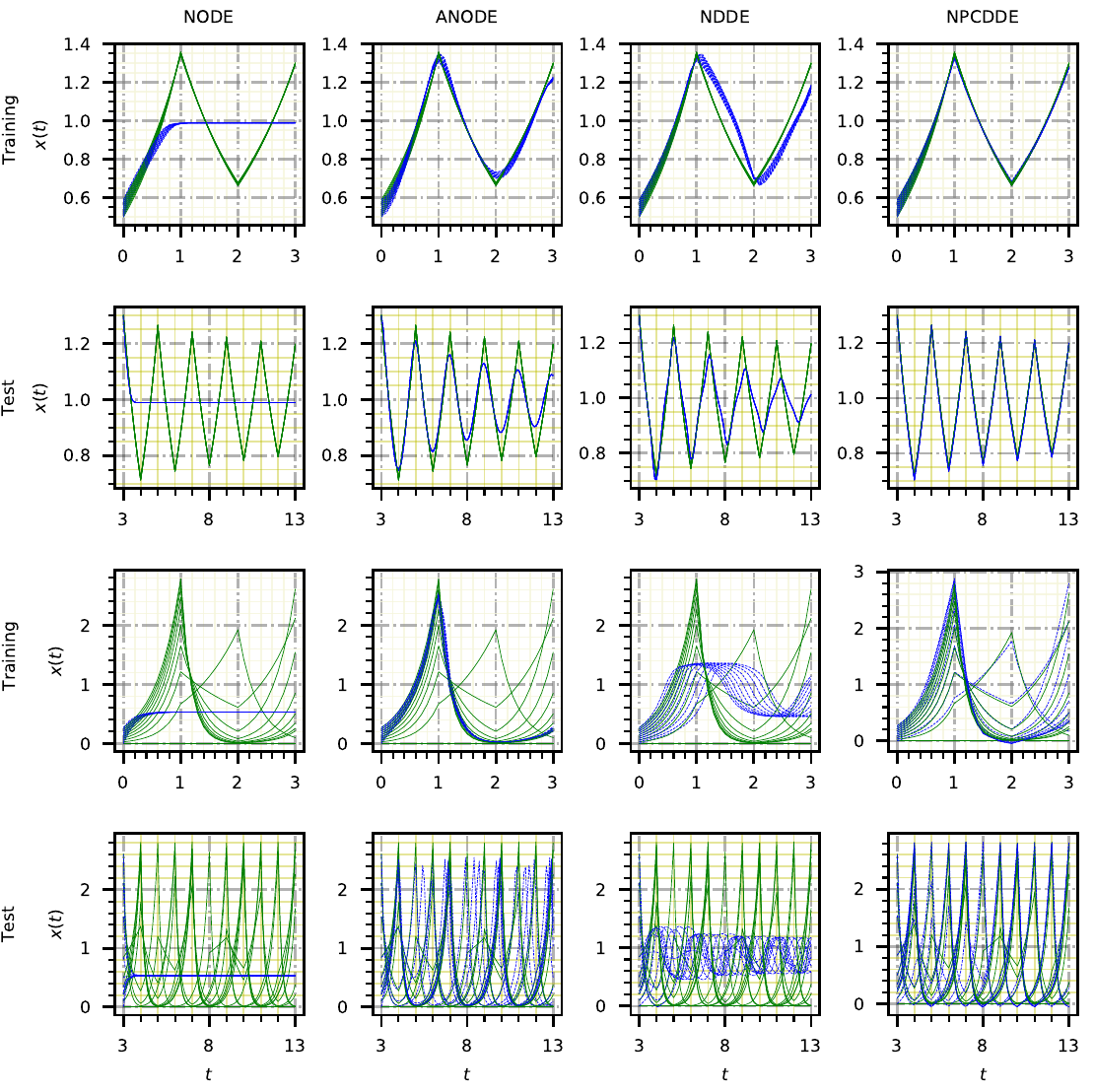}
        \end{center}
        \caption{
        {The trajectories of the piecewise-constant population dynamics exhibiting oscillating (the first two rows) and chaotic (the last two rows) dynamics in the training and the test stages. Here, displayed are $10$ trajectories from the total $100$ trajectories for the training and the testing time-series. The solid lines correspond to the true dynamics, while the dash lines correspond to the trajectories generated by the trained models in the training duration and in the test duration, respectively.}
        }
        \label{nfe_forward_fig}
    \end{figure*}

    \begin{figure*}[htb]
    \begin{center}
        \includegraphics[width=13.3cm]{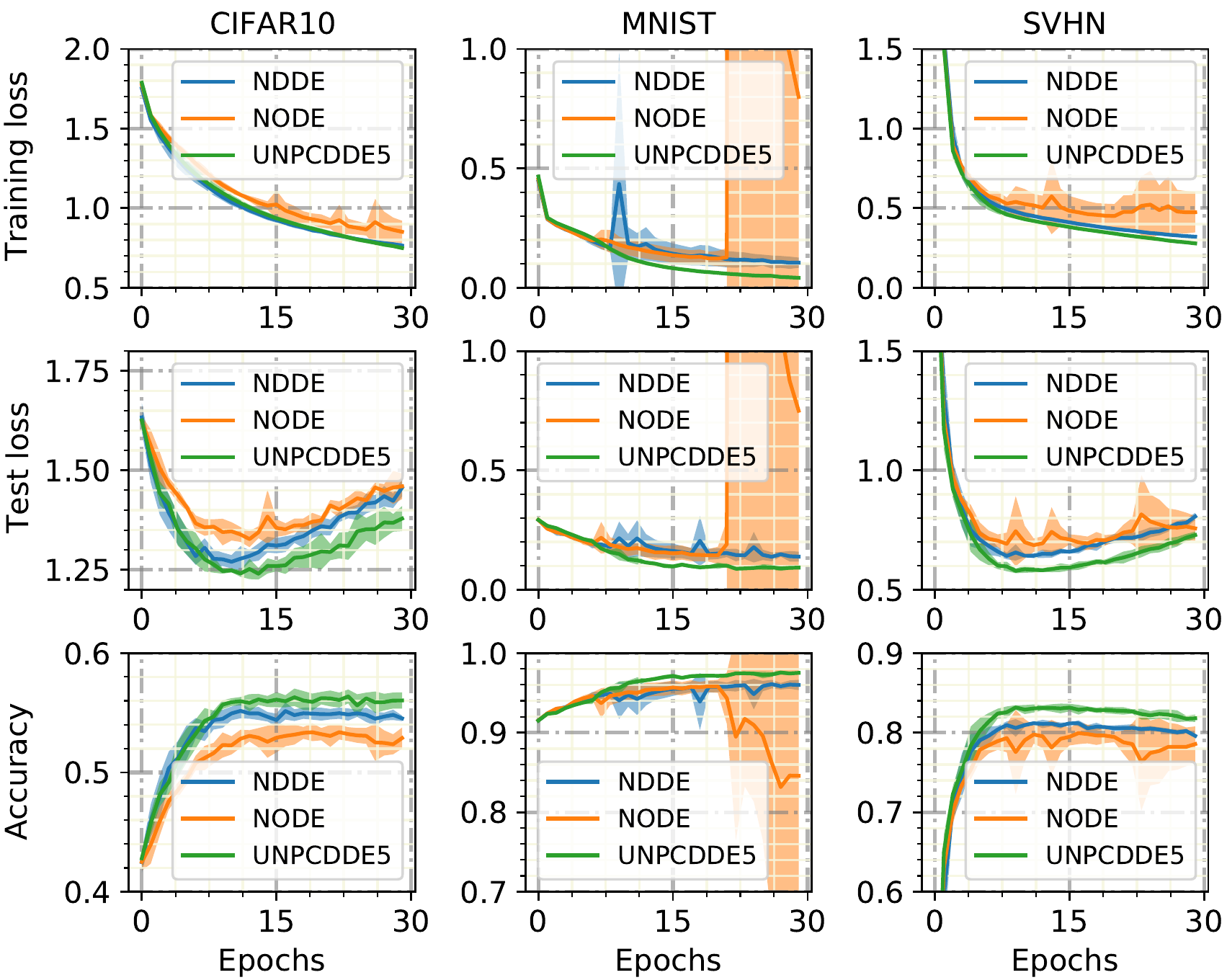}
    \end{center}
    \caption{The training losses (top row), the test losses (middle row), and the accuracies (bottom row), respectively, by using the NODEs, the NDDEs, the UNPCDDEs in \eqref{eqDNDDEgenralunshared} with $n=5$ over $5$ realizations for the image datasets, including CIFAR10 (left), MNIST (middle), and SVHN (right). Here, 
    the panels for the NODEs and the NDDEs are directly adapted from \citep{zhu2021neural}. The final time $T$ for all models equals to $1$.
    }
    \label{mnist_cifar_fig}
    \vskip -0.1in
    \end{figure*}

    \begin{figure*}[htb]
        \begin{center}
            \includegraphics[width=17cm]{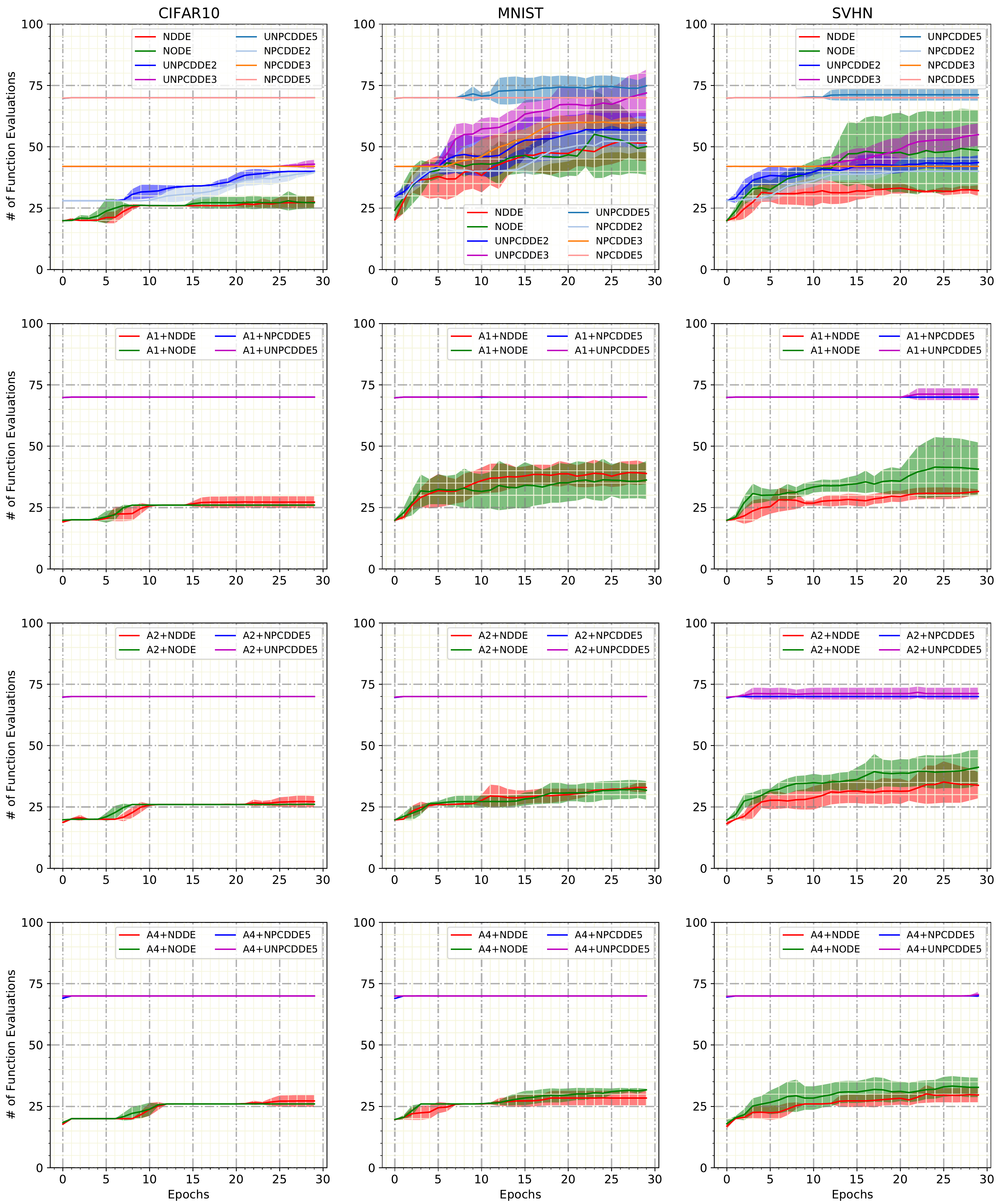}
        \end{center}
        \caption{
        {Evolution of the forward number of the function evaluations (NFEs) during the training process for each model on the image datesets: CIFAR10, MNIST, and SVHN.}
        }
        \label{nfe_forward_fig}
    \end{figure*}
    
    \begin{figure*}[htb]
        \begin{center}
            \includegraphics[width=17cm]{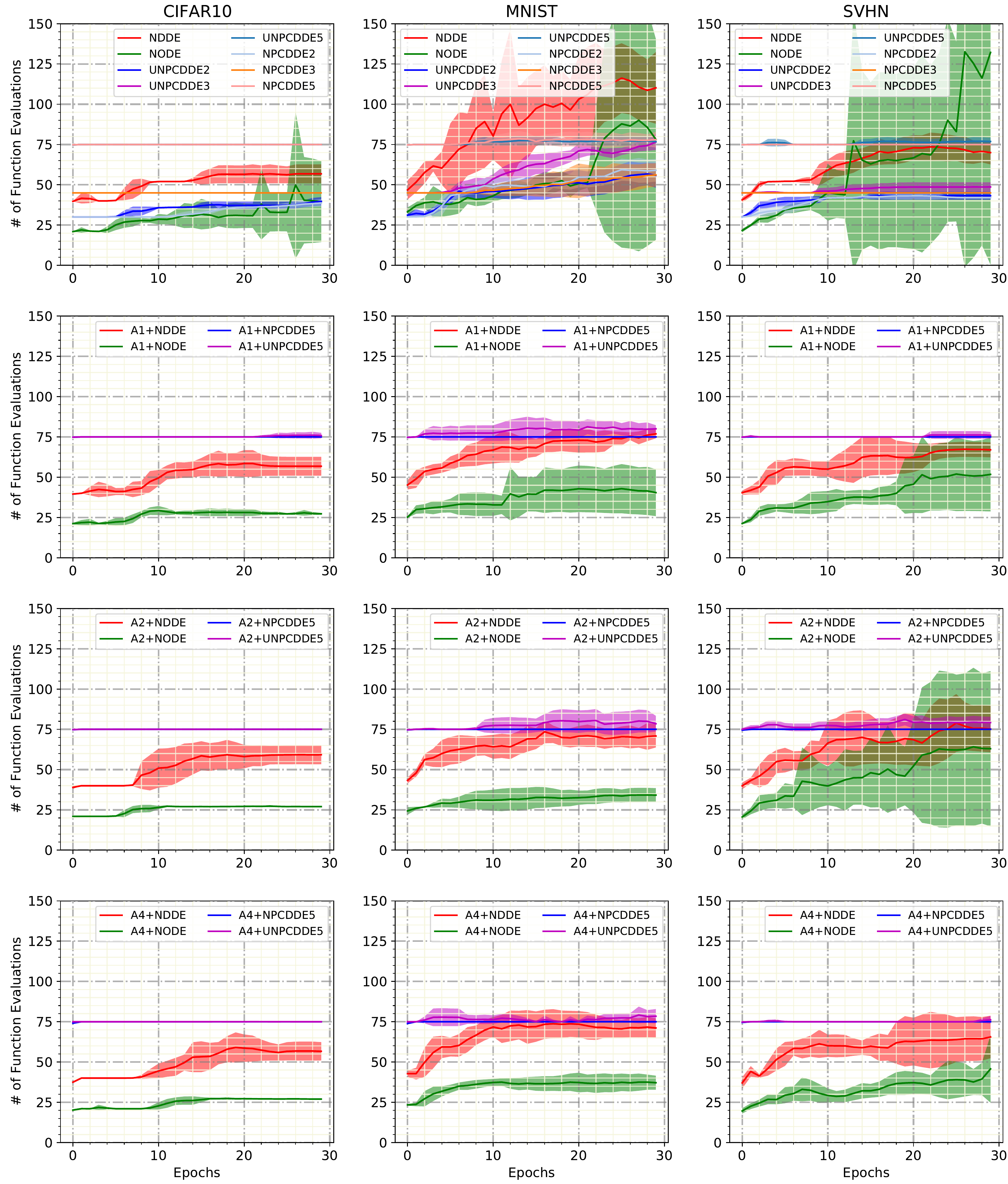}
        \end{center}
        \caption{
        {Evolution of the forward number of the function evaluations (NFEs) during the training process for each model on the image datesets: CIFAR10, MNIST, and SVHN.}
        }
        \label{nfe_forward_fig}
    \end{figure*}
    
\end{document}